\documentclass[lettersize,journal]{IEEEtran}
\usepackage{caption}
\usepackage{amsmath,amsfonts}
\usepackage{algorithm}
\usepackage{tabularray}
\usepackage{color}
\usepackage{array}
\usepackage[caption=false,font=normalsize,labelfont=sf,textfont=sf]{subfig}
\usepackage{textcomp}
\usepackage{stfloats}
\usepackage{float}
\usepackage{url}
\usepackage{verbatim}
\usepackage{graphicx}
\usepackage{cite}
\usepackage{subfig}
\usepackage{subfloat}
\usepackage{amsmath}
\usepackage[caption=false,font=normalsize,labelfont=sf,textfont=sf]{subfig}
\usepackage{caption}
\usepackage{subcaption}
\usepackage{makecell}
\usepackage{booktabs}
\usepackage{multirow}
\usepackage{algorithmicx}
\usepackage[noend]{algpseudocode}
\usepackage[table,xcdraw]{xcolor}
\usepackage[table,xcdraw]{xcolor}
\usepackage{color}
\usepackage{lineno}
\usepackage{nomencl}
\usepackage{amsmath}
\usepackage{hyperref} 

\makenomenclature

\begin{document}
\graphicspath{ {./picture/} }
\bibliographystyle{IEEEtran}

\title{Task-Oriented Real-time Visual Inference for IoVT Systems: A Co-design Framework of Neural Networks and Edge Deployment}
\author{Jiaqi Wu, Simin Chen, Zehua Wang, Wei Chen, Zijian Tian, \textcolor{black}{F. Richard Yu,\textit{ Fellow IEEE}}, Victor C. M. Leung, \textit{life Fellow IEEE}
\thanks{
}

\thanks{\textit{This work has been submitted to the IEEE for possible publication. Copyright may be transferred without notice, after which this version may no longer be accessible.} ({\emph{Corresponding author: Simin Chen, Wei Chen.}})
}

\thanks{Jiaqi Wu is with the School of Artificial Intelligence, China University of Mining and Technology (Beijing), Beijing 100083, China. The author is also with Department of Electrical and Computer Engineering, University of British Columbia, Vancouver, 2332 Main Mall Vancouver, BC Canada V6T 1Z4 (e-mail: wjq11346@student.ubc.ca)
}

\thanks{Simin Chen is with the university of Texas at dallas, 800 W Campbell Rd, Richardson, TX 75080. (e-mail:sxc180080@utdallas.edu)
}

\thanks{Zijian Tian is with the School of Artificial Intelligence, China University of Mining and Technology (Beijing), Beijing 100083, China.(e-mail:Tianzj0726@126.com)
}

\thanks{Wei Chen is with the Ministry of Emergency Management Big Data Center, Beijing 100013, China. The author is also with School of Artificial Intelligence, China University of Mining and Technology (Beijing), Beijing 100083, China. (e-mail: chenwdavior@163.com)
}

\thanks{Zehua Wang is with the Department of Electrical and Computer Engineering, The University of British Columbia, Vancouver, 2332 Main Mall Vancouver, BC Canada V6T 1Z4 (e-mail: zwang@ece.ubc.ca).}



}

\markboth{IEEE Journal on Selected Areas in Communication}%
{Shell \MakeLowercase{\textit{et al.}}: A Sample Article Using IEEEtran.cls for IEEE Journals}


\maketitle


\begin{abstract}

As the volume of collected video data continues to grow, data-oriented cloud computing in Internet of Video Things (IoVT) systems faces increasing challenges, such as system latency and bandwidth pressure. Task-oriented edge computing, by shifting data analysis to the edge, enables real-time visual inference. However, the limited computational power of edge devices presents challenges for effectively executing computationally intensive visual tasks. Existing methods struggle with the trade-off between high model performance and low resource consumption. For instance, lightweight neural networks often deliver limited performance, while Neural Architecture Search (NAS)-based model design incurs high computational and commercial costs at training Inspired by hardware/software co-design principles, we propose, for the first time, a co-design framework at both the model and system levels to optimize neural network architecture and deployment strategies during inference. This framework maximizes the computational efficiency of edge devices to achieve high throughput, enabling real-time edge inference. Specifically, we implement a dynamic model structure based on the re-parameterization principle, coupled with a Roofline-based model partitioning strategy to synergistically enhance the computational performance of heterogeneous devices. Furthermore, we employ a multi-objective co-optimization approach for the above strategies to balance throughput and accuracy. We also conduct a thorough analysis of the mathematical consistency between the partitioned and original models, in addition to deriving the convergence of the partitioned model. Experimental results show that, compared to baseline algorithms, our method significantly improves throughput (12.05\% on MNIST, 18.83\% on ImageNet) while also achieving higher classification accuracy. Additionally, it ensures stable performance across devices with varying computational capacities, highlighting its generalizability and practical application value. Simulated experiments further validate that our method enables high-accuracy real-time detection in IoVT systems, particularly for small object detection.

\end{abstract}





\section{Introduction}

\begin{figure}[t]
    \centering
    \includegraphics[width=\linewidth]{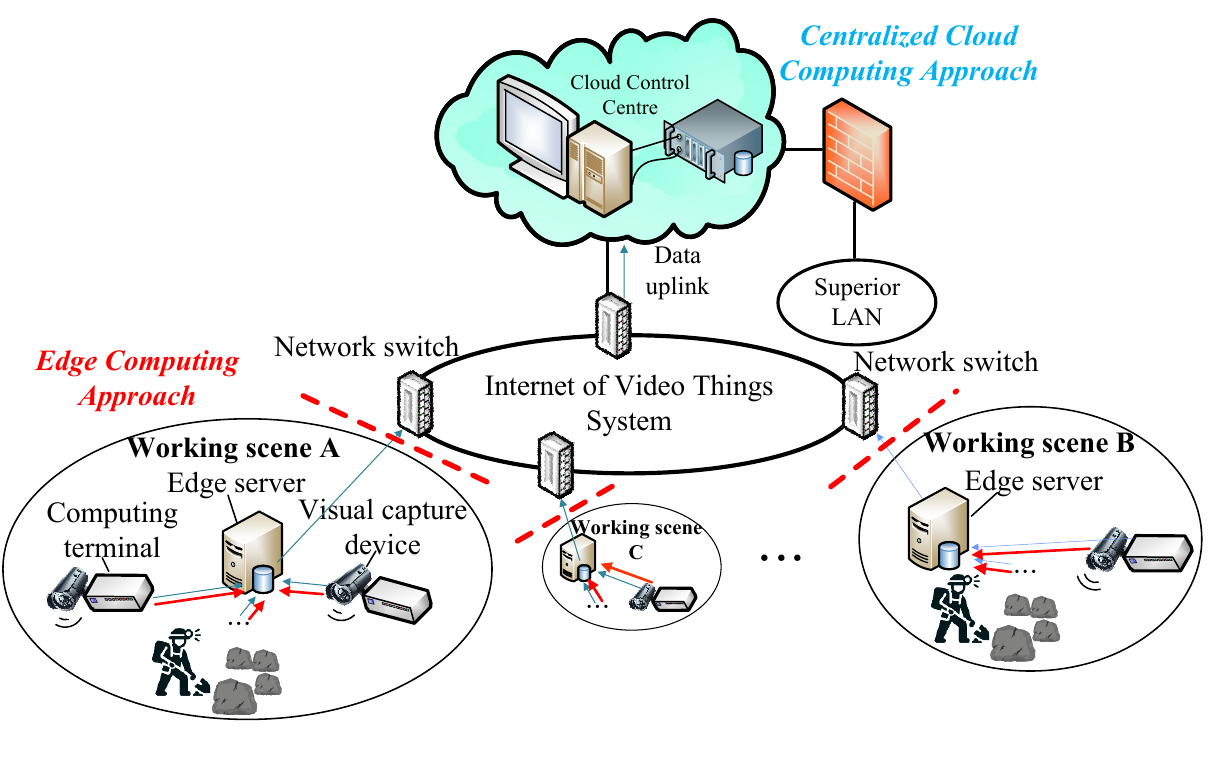}
    \caption{The comparison between centralized cloud computing and edge computing approach in the IoVT system. Centralized cloud computing is a typical data-driven design, requiring all data to be uploaded to a central server for inference. In contrast, edge computing is a task-oriented paradigm, where data analysis tasks are offloaded from the cloud to the edge, closer to the data collection points, in order to reduce system latency.}
    \label{fig:cloud-edge}
\end{figure}

Real-time visual task inference is a core function of the Internet of Visual Things (IoVT) systems\cite{chen2020internet,anand2022smart}, particularly for time-sensitive tasks such as object detection\cite{liang2022edge} and object tracking\cite{shao2019enabling}. Real-time inference can effectively prevent task failures caused by latency. Currently, many IoVT systems rely on a centralized cloud computing architecture, as shown in Fig. \ref{fig:cloud-edge}, where data collected by terminal sensing devices is transmitted to cloud servers for centralized inference. However, as the number of terminal devices increases and the volume of visual data grows exponentially, the long-distance transmission of large-scale data creates immense bandwidth pressure, severely compromising the system’s real-time performance\cite{guo2021noma,wu2024lightweight}. Thus, this centralized cloud computing design principle is not well-suited for machine vision applications.

Edge computing\cite{chen2019deep}, as an emerging computational paradigm, as illustrated in Fig. \ref{fig:cloud-edge}, shifts data processing and analysis tasks closer to the data source, avoiding the need for massive data transmission, and thus holds greater potential for enhancing real-time system performance.\cite{liu2019survey} For example, studies \cite{wu2024lightweight} and \cite{zhu2022real} deploy lightweight object detection models on edge devices (e.g., drones) to achieve real-time power inspection, while study \cite{li2019edge} deploys deep learning models on edge servers to avoid the latency issues associated with transmitting visual data to the cloud. 
However, a key challenge is limited computational resources of each edge device, which are insufficient to meet high-performance demands of complex visual tasks\cite{shuvo2022efficient}. Although the computational power of edge devices has improved, it still lags significantly behind cloud servers, making it difficult to fully satisfy the computational requirements of visual inference tasks \cite{lim2024cutting,zhang2024dvfo}.



Despite neural network edge deployment solutions being proposed to address this issue\cite{cao2021large, wang2019joint}, there remain significant limitations in achieving real-time inference, where balancing performance and throughput proves challenging\cite{cruz2022edge}. To explore this, we introduce several key edge deployment approaches and analyze their drawbacks:

\begin{itemize}
    \item \textbf{General lightweight models} (e.g., the Mobilenet family\cite{howard2019searching} and GhostNet\cite{liu2024ghostnetv3}): These models focus on lightweight design to reduce computational complexity, making them suitable for embedded devices with limited resources\cite{chen2020deep}. However, despite reducing computational load, their feature representation capabilities are limited, resulting in \textit{suboptimal performance for high-accuracy visual inference tasks}.
    
    \item \textbf{Neural Architecture Search (NAS)-based methods}: These methods employ automated network structure search strategies to design models optimized for specific edge devices\cite{zheng2023ddpnas, liu2021fox}. While theoretically capable of customizing models for device-specific deployment, the process incurs significant computational and financial costs\cite{ren2021comprehensive}. Moreover, NAS-generated models are typically device-specific, \textit{lacking generalizability across devices, which severely limits their practical application value.}
    
    \item \textbf{Split learning approaches}: Split learning methods \cite{thapa2022splitfed} partition large-scale models across different devices for distributed sequential inference. Studies \cite{wu2024lightweight} and \cite{xu2023learning} introduce a split learning strategy for model distribution deployment in industrial IoVT systems to achieve real-time visual inference. Study \cite{roofsplit} proposes a model partitioning deployment method based on roofline analysis to enhance inference throughput. However, relying solely on the model partitioning strategy still has \textit{limited effectiveness in maximizing the device's computational performance to improve throughput.}
\end{itemize}


To address these issues, we propose, for the first time, a co-design method for neural network architecture and edge deployment tailored for real-time visual inference in IoVT systems. This approach performs multi-objective joint optimization of dynamic model architecture design and deployment strategies to enhance the compatibility between the model and edge devices, maximizing computational power and improving throughput while ensuring high accuracy. At the system level, we adopt a Roofline-based\cite{williams2009roofline} model partitioning strategy that determines the optimal partitioning scheme based on the peak computational performance and memory bandwidth of the hardware, fully leveraging their computational resources to boost throughput. At the model level, we employ a dynamic architecture based on a reparameterization strategy\cite{ding2021repvgg}, which adaptively adjusts the neural network during inference to reduce computational complexity while maintaining model performance, further optimizing throughput. To simultaneously enhance throughput and maintain model accuracy, we utilize a multi-objective co-optimization approach that coordinates both model architecture and deployment strategies. Unlike traditional optimization methods, our approach dynamically adjusts the pre-trained model during inference, enabling it to adapt to different devices with greater generalizability, making it more suitable for IoVT systems where frequent dynamic device participation occurs. Experimental results demonstrate that, compared to six state-of-the-art (SOTA) methods, the proposed approach significantly improves throughput by an average of 12.05\% on MNIST, 13.75\% on CIFAR-100, and 18.83\% on ImageNet, while achieving satisfactory classification accuracy. Additionally, it achieves excellent throughput and accuracy across different devices, highlighting its generalizability. Furthermore, we evaluated the practical application of the proposed method in a simulated industrial IoVT system for object detection. The results show that our approach enables successful edge-side model deployment, performing real-time and accurate small object detection. The main contributions of this paper are summarized as follows:

\begin{itemize}
    \item We propose a co-design strategy for neural network architecture and partitioned deployment to maximize edge devices' computing performance in IoVT systems. This strategy significantly improves system throughput to enables real-time edge inference.

    \item A multi-objective co-optimization approach is adopted to effectively improve throughput while maintaining model performance. 
    
    \item The co-design process is performed during the application phase, adjusting the pre-trained model to adapt to different devices. This applies to practical IoVT systems with varying device registration and deregistration scenarios.

    \item Experimental results demonstrate that the proposed method effectively enhances edge inference speed while preserving model performance. The co-design strategy at both the model and system levels proves to be highly practical.
\end{itemize}


\section{Related Work}
\label{Related Work}
\begin{table*}[t]
\setlength{\tabcolsep}{0.08cm} 
\centering
\caption{Comparison of Co-design Algorithms. }
\begin{tabular}{lcccccccc}
\toprule
\textbf{Work}     & \makecell{\textbf{Hardware-Software} \\ \textbf{Co-design}} & \makecell{\textbf{Model-level} \\ \textbf{Design}} & \makecell{\textbf{Edge} \\ \textbf{Deployment}} & \makecell{\textbf{Memory} \\ \textbf{Efficiency}} & \makecell{\textbf{Latency} \\ \textbf{Reduction}} & \makecell{\textbf{Performance} \\ \textbf{Improvement}} & \makecell{\textbf{Task-Specific} \\ \textbf{Optimization}} & \makecell{\textbf{Optimization} \\ \textbf{Objectives}} \\ \midrule
\textbf{Neo\cite{mudigere2022software}}      & \checkmark                           & \checkmark                        & $\times$                           & \checkmark                 & $\times$                           & \checkmark                       & DLRM Training                      & Accuracy, Efficiency             \\
\textbf{Prodigy\cite{talati2021prodigy}}  & \checkmark                           & $\times$                                  & $\times$                           & \checkmark                 & $\times$                           & \checkmark                       & Irregular Workloads                & Latency, Energy                  \\
\textbf{SECDA\cite{haris2021secda}}    & \checkmark                           & $\times$                                  & \checkmark                & \checkmark                 & \checkmark                & \checkmark                       & DNN Acceleration on Edge           & Speedup, Energy                  \\
\textbf{SDP\cite{sdp}}      & \checkmark                           & $\times$                                  & $\times$                           & \checkmark                 & $\times$                           & \checkmark                       & Sparse Neural Network              & Area Efficiency, Energy          \\
\textbf{ECSSD\cite{li2023ecssd}}    & \checkmark                           & $\times$                                  & $\times$                           & \checkmark                 & $\times$                           & $\times$                                   & Extreme Classification             & Performance                      \\
\textbf{ASH\cite{elsabbagh2023accelerating}}      & \checkmark                           & $\times$                                  & $\times$                           & \checkmark                 & $\times$                           & \checkmark                       & RTL Simulation                     & Speedup, Area                    \\
\textbf{MP-Rec\cite{hsia2023mp}}   & \checkmark                           & $\times$                                  & $\times$                           & \checkmark                 & $\times$                           & \checkmark                       & Embedding Representation           & Efficiency, Speedup              \\
\textbf{xURLLC\cite{park2020extreme}}   & \checkmark                           & $\times$                                  & $\times$                           & N/A                        & $\times$                           & $\times$                                   & 5G URLLC                           & Reliability, Latency             \\
\textbf{HAAC\cite{haac}}   & \checkmark                           & $\times$                                  & $\times$                           & \checkmark                 & $\times$                           & \checkmark                       & Neural Network Attention           & Scalability, Efficiency           \\
\textbf{Our method} & \checkmark                         & \checkmark                        & \checkmark                & \checkmark                 & \checkmark                & \checkmark                       & Deep Learning in IoVT              & Latency, Performance             \\ \bottomrule
\end{tabular}
\label{tab:codesignsummary}
\end{table*}
\subsection{Co-design mechanism for computing efficiency}

Co-design mechanism\cite{zhang2022algorithm} is widely applied to reduce computation latency, focusing on joint design across different layers, such as hardware and software levels, to fully leverage hardware capabilities and minimize system delays. This approach enables a more efficient balance between algorithmic complexity and hardware performance, making it a key strategy in many modern computational tasks\cite{hao2019fpga}. Below are examples of various co-design algorithms that illustrate these principles:
Neo\cite{mudigere2022software} is a co-designed software-hardware system that efficiently trains large-scale deep learning recommendation models (DLRMs) by using 4D parallelism to optimize embedding computations. Paired with the ZionEX platform, Neo improves DLRM training performance through system optimizations like hybrid kernel fusion and caching.
Prodigy\cite{talati2021prodigy} is a hardware-software co-designed prefetching solution that improves memory latency for irregular workloads by leveraging static program information and dynamic run-time data. By utilizing the Data Indirection Graph (DIG) to guide hardware prefetching, Prodigy enhances both performance and energy efficiency.
SECDA\cite{haris2021secda} is a co-design methodology that accelerates the development of optimized DNN accelerators on edge devices with FPGAs by combining SystemC simulation with hardware execution, reducing design time and complexity. This approach facilitates efficient design space exploration and improves performance and energy savings over CPU-based inference.
SDP\cite{sdp} is a co-designed digital Processing-in-Memory (PIM) architecture for sparse neural network acceleration, integrating algorithm, dataflow, and hardware optimizations. By employing hybrid-grained pruning and bit-serial Booth multiplication, SDP improves area efficiency and energy savings over other sparse NN architectures.
ECSSD\cite{li2023ecssd} is a hardware and data layout co-designed in-storage-computing architecture for extreme classification, optimizing SSD-based large-scale workloads. By combining an alignment-free MAC circuit and a heterogeneous data layout, ECSSD significantly enhances performance over existing baselines.
ASH\cite{elsabbagh2023accelerating} is a hardware-software co-designed architecture for fast RTL simulation, combining a specialized hardware architecture with a compiler optimized for parallel execution. By using dataflow execution and selective event-driven processing, ASH achieves significant speedup over traditional simulators while using less area.
MP-Rec\cite{hsia2023mp} is a hardware-software co-design technique that optimizes embedding representations for deep learning recommendation systems. By leveraging custom accelerators such as GPUs and TPUs, MP-Rec enhances memory efficiency and system performance for large-scale recommendation tasks.
xURLLC\cite{park2020extreme} builds on 5G URLLC by addressing stricter demands in reliability, latency, and scalability through machine learning, non-radiofrequency data integration, and joint communication-control co-design. These innovations aim to overcome the limitations of 5G URLLC for emerging high-stakes applications.
This literature\cite{haac} proposes a hardware-software co-designed accelerator and compiler (HAAC) that transforms garbled circuits (GCs) into data streams, simplifying hardware design and decoupling memory and computation. The approach leverages the compiler's program comprehension to significantly enhance the performance and efficiency of privacy-preserving computation, while maintaining system generality.
To reduce system latency of deep learning models in IoVT systems, we propose a model-level and system-level co-design approach. This method performs a coordinated optimization of neural
networks structure and model partition deployment to enhance the adaptability between partitioned modules and hardware devices, thereby maximizing the computational performance of resource-constrained edge devices and reducing latency. We summarize the aforementioned methods in TABLE \ref{tab:codesignsummary}, where it is evident that our method demonstrates more comprehensive performance.

\subsection{Edge deployment model designing}
Edge computing\cite{chen2019deep} is an emerging paradigm that shifts data inference tasks from cloud servers to the edge of IoT systems, mitigating the latency caused by massive data transmission. Designing models for edge deployment is a significant challenge, as it requires models to be lightweight to fit resource-constrained edge devices while maintaining high performance. Existing solutions include:
\begin{itemize}
    \item Utilizing the Neural Architecture Search (NAS) strategies to design device-specific network architectures based on hardware capabilities.
    \item Developing general-purpose lightweight neural networks.
\end{itemize}

\textit{NAS-based methods}: OnceNAS\cite{zhang2024oncenas} introduces a NAS method that simultaneously optimizes parameter count, inference latency, and accuracy for edge devices, using dynamic evaluation to adapt networks to hardware limitations. DGL-based NAS\cite{dgl} predicts network latency on new devices without direct sampling, enabling efficient architecture searches by modeling device parameters. Loong\cite{loong} employs a layer-wise NAS approach, embedding and training candidate operations with gradient feedback, while customizing architectures for heterogeneous devices by optimizing memory and latency. DLW-NAS\cite{li2023dlw} uses a differentiable search space focused on lightweight operations, integrating complexity constraints to optimize networks for resource-limited devices.

\begin{figure}[h]
    \centering
    \includegraphics[width=9cm, height=6cm]{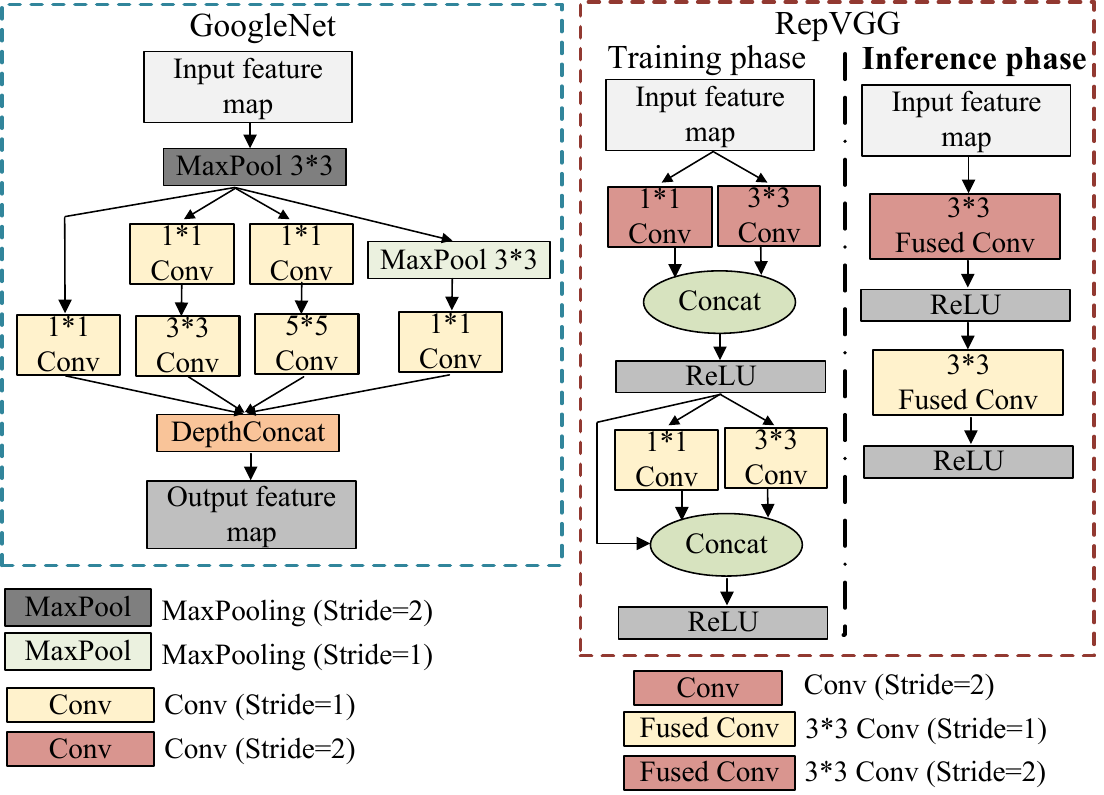}
    \caption{Dynamic model architecture of RepVGG. GoogLeNet uses convolutional layers with different kernel sizes to extract multi-scale features, but this also significantly increases the model's computational complexity. RepVGG, on the other hand, utilizes multi-branch structures during the training phase to extract rich features and merges these branches during the inference phase. This approach reduces the model size while effectively maintaining performance.}
    \label{fig:dynamicmodelstructure}
\end{figure}

\textit{Lightweight neural networks}: In recent years, several lightweight network architectures have been developed to improve computational efficiency while maintaining accuracy. MobileNetV1\cite{howard2017mobilenets} utilized Depthwise Separable Convolutions and introduced hyperparameters $\alpha$ and $\beta$ to control convolution kernel size and input image resolution, significantly reducing computational costs. EfficientNetV1\cite{tan2019efficientnet} introduced a compound scaling method to balance network width, depth, and resolution for optimal efficiency. EfficientNetV2\cite{tan2021efficientnetv2} further enhanced this with a combination of training-aware neural architecture search and Fused-MBConv modules to optimize performance on modern accelerators. Based on the split-transform-merge structure of GoogLeNet\cite{module2015googlenet}, as shown in Fig. \ref{fig:dynamicmodelstructure}, RepVGG\cite{ding2021repvgg} proposes a novel solution by decoupling the training and inference architectures, allowing the model to benefit from multi-branch training while simplifying the structure for deployment, as illustrated in Fig. \ref{fig:dynamicmodelstructure}. 

Inspired by RepVGG, we propose a dynamic model structure based on re-parameterization, which adjusts the network architecture during inference to match the performance of the current computing device, thereby maximizing its computational capabilities and improving inference throughput. Unlike NAS-based methods that design device-specific networks, this approach adaptively adjusts the model's architecture based on the device's performance, offering greater practical applicability.

\section{Methodology}
\label{Methodology}

\subsection{Overview}
\label{Decentralized Coal Mine Video Surveillance System}

\begin{figure*}[t]
    \centering
    \fbox{\includegraphics[width=17cm, height=7cm]{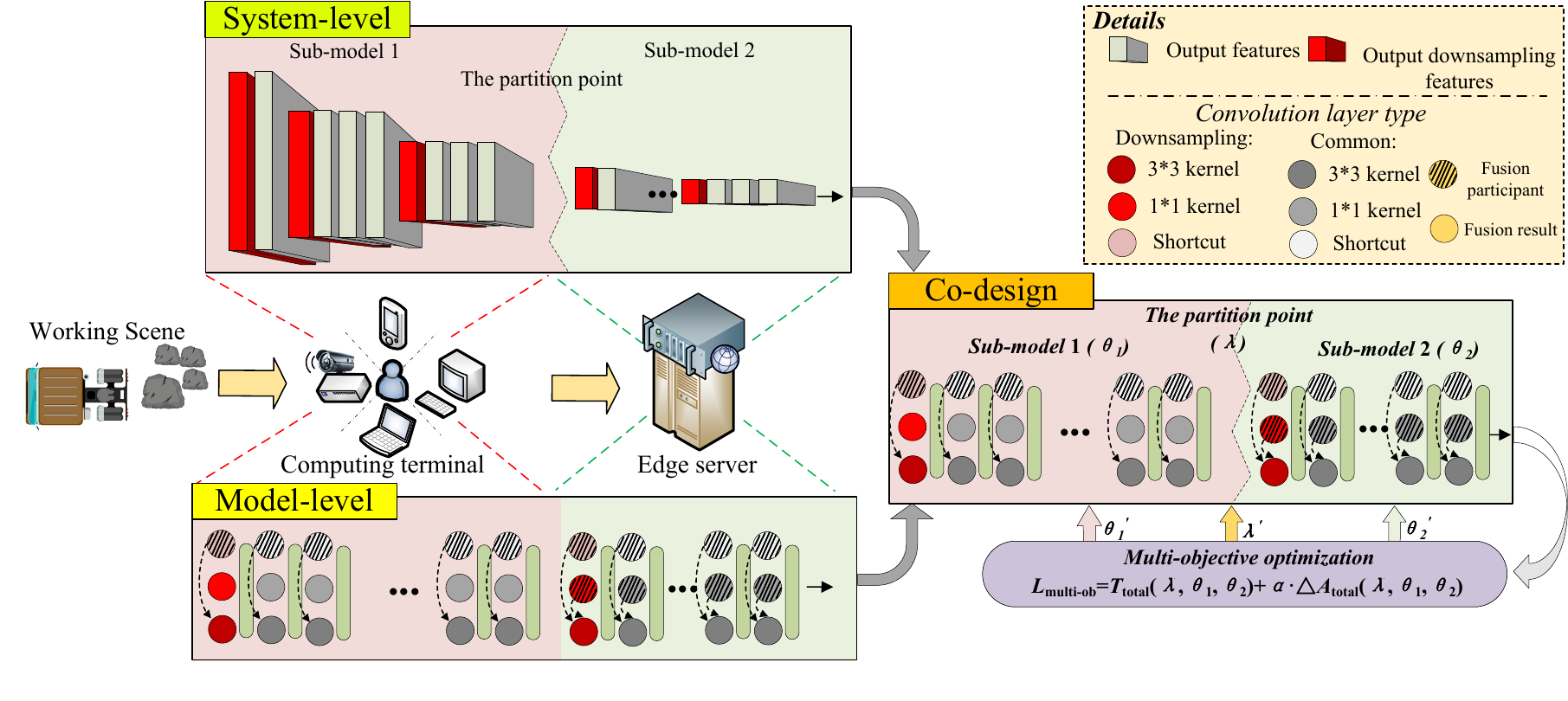}}
    \caption{The overview of our method. At the \textbf{system-level}, we employ a model partitioning method based on Roofline analysis, where sub-model 1 and sub-model 2 are deployed on the computing terminal and the edge server, respectively, according to the selected partition point. At the \textbf{model-level}, we utilize a dynamic model structure that adapts the network architecture during the inference phase to better match the computational resources. \textbf{Multi-objective optimization}, as shown in Eq. 11, adjusts both the partition point and model structure to fully leverage the computational performance of the devices, thereby improving throughput while maintaining inference accuracy.}
    \label{fig:overview}
\end{figure*}

We propose a co-design method for model dynamic structure and edge partition deployment tailored for real-time inference on resource-constrained edge devices in IoVT systems.\footnote{https://github.com/word-ky/Co-design-Sys-NN} The architecture of the method is shown in Fig. \ref{fig:overview}. At the system level, we utilize Roofline analysis (see Fig. \ref{fig:roofline}) to determine the partition point by matching the partitioned model with the corresponding device’s capabilities, thereby maximizing the device’s computational performance. At the model level, we adopt a dynamic model structure based on re-parameterization, as depicted in Fig. \ref{fig:dynamicmodelstructure}, and employ adaptive network structure to further optimize model-device alignment. This approach uses dynamic network structures to facilitate model partitioning, enhancing the flexibility and practicality of designing model-hardware matching solutions. Additionally, we incorporate a multi-objective optimization function with the hyperparameter $\lambda_1$ to perform system-model co-design, significantly reducing system latency while effectively maintaining model accuracy.
         
\section{neural network/edge deployment co-design method}

We propose a neural network / edge deployment strategies co-design approach. Next, we provide a detailed introduction to the system-level edge partitioning deployment strategy, the neural network-level dynamic model structure, and the multi-objective co-optimization approach.
We propose a neural network / edge deployment strategies co-design approach. Next, we provide a detailed introduction to the system-level edge partitioning deployment strategy, the neural network-level dynamic model structure, and the multi-objective co-optimization approach.

\textit{Computational Intensity Requirements}. To ensure that the partitioned sub-models operate efficiently on their respective hardware platforms, they must satisfy the computational intensity requirements of each sub-model. The Roofline analysis\cite{williams2009roofline} helps to fully utilize hardware resources, preventing system bottlenecks caused by limitations in memory bandwidth or computational capacity. Computational intensity (Operational Intensity, \( I \)) is defined as the ratio of floating-point operations (FLOPs) to data transfer volume (in bytes), measured in FLOP/byte. The concept of computational intensity reflects the balance between the system's computational tasks and its data transfer requirements. Each hardware platform has its own computational intensity threshold: \( I_m = \frac{\pi}{\beta} \), where \( \pi \) represents the peak computational capability of the hardware (in FLOP/s), and \( \beta \) represents the memory bandwidth (in bytes/s). The computational intensity threshold (in FLOP/byte) is used to evaluate the hardware’s ability to execute tasks efficiently.

\begin{figure}[t]
    \centering
    \includegraphics[width=\linewidth]{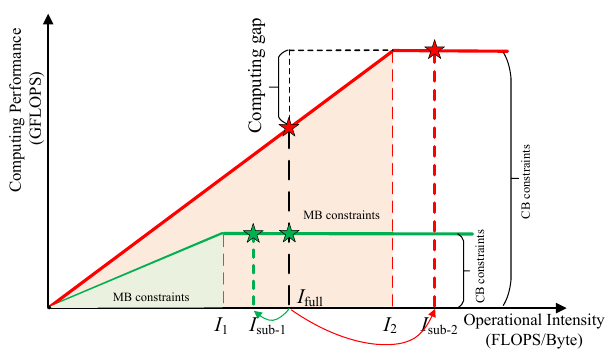}
    \caption{The Roofline results of the partitioned model. It shows that $\textit{I}_{\text{mn}}$ represents the computational intensity of different devices. The computational intensity of the model, $\textit{I}_0$, is smaller than $\textit{I}_{m2}$, which indicates that it is constrained by the memory bandwidth (MB constraints) of Device 2, and larger than $\textit{I}_{m1}$, indicating it is constrained by the computational performance (CB constraints) of Device 1. The goal of segmentation deployment is that, after partitioning $\textit{I}_0$, the computational intensity of the segmentation modules, $\textit{I}_1$ and $\textit{I}_2$, should optimally leverage the maximum computational performance of the corresponding devices.
}
    \label{fig:roofline}
\end{figure}

The computational intensity of the complete model is expressed as \( I_0 = \frac{\sum_i C_i}{\sum_i M_i} \). \( C_i \) represents the floating-point operations (FLOPs) of sub-model $i$, while \( M_i \) denotes the memory size in bytes required for a single feed-forward propagation. In this study, we partition the model into two sub-models, where the computational intensity of each sub-model is defined as \( I_i = \frac{C_i}{M_i} \), with \( i \in \{1, 2\} \). The Roofline results of model partitioning processing as shown in Fig. \ref{fig:roofline}. The computational intensity of sub-model 1, namely \( I_1 \), representing the intensity of the sub-model on the computing terminal, is given by the following equation:

\begin{equation}
I_1 = \frac{C_1}{M_1} = \frac{\lambda_C C_0}{\lambda_M M_0} = \frac{\lambda_C}{\lambda_M} I_{0} ,
\end{equation}
where the parameter \( \lambda \) represents the partition point, where \( \lambda_C \) and \( \lambda_M \) indicate the proportion of the model partitioned in terms of floating-point operations and memory size, respectively.

To ensure that sub-model 1 operates efficiently, the computational intensity \( I_1 \) should ideally be as large as possible to maximize the computational performance of the device. If \( I_1 \) reaches or exceeds the computational intensity threshold \( I_{m1} \) of hardware 1, the device will be able to achieve its peak performance. This relationship can be expressed as:

\begin{equation}
\frac{\lambda_C}{\lambda_M} \cdot I_0 \geq I_{m1} = \frac{\pi_1}{\beta_1}.
\end{equation}
The computational intensity of sub-model 2 \( I_2 \), representing the intensity of the sub-model on the edge server, is given by the following equation:
\begin{equation}
I_2 = \frac{(1 - \lambda_C)}{(1 - \lambda_M)} \cdot I_0.
\end{equation}
If \( I_2 \) is greater than or equal to the computational intensity threshold \( I_{m2} \) of hardware 2, the edge server will operate efficiently. This condition is expressed as:
\begin{equation}
\frac{(1 - \lambda_C)}{(1 - \lambda_M)} \cdot I_0 \geq I_{m2} = \frac{\pi_2}{\beta_2}.
\end{equation}
By satisfying these computational intensity requirements, each sub-model can efficiently utilize the resources of its respective hardware platform, avoiding system bottlenecks caused by limited memory bandwidth or computational capacity.

\begin{algorithm}
\caption{System Latency Optimization for Partitioned Sub-models}
\label{alg:latency_optimization}
\begin{algorithmic}[1]
\Statex \textbf{Input:} Initial partition point $\lambda_0$, initial computational intensity $I_0$
\Statex \textbf{Output:} Optimized partition point $\lambda^*$, total system latency $T_{\text{total}}^*$

\State Initialize partition point $\lambda \leftarrow \lambda_0$
\For{$\text{iteration} = 1$ to $\text{max\_iterations}$}
    
    \State \textbf{Step 1: Calculate Computational Intensities}
    \State $I_1 \leftarrow \frac{\lambda_C}{\lambda_M} \cdot I_0$, $I_2 \leftarrow \frac{(1 - \lambda_C)}{(1 - \lambda_M)} \cdot I_0$
    
    \State \textbf{Step 2: Compute Sub-model Latencies}
    \State $T_1 \leftarrow \frac{\lambda N}{\mu_{11} \cdot I_1 \cdot \beta_1}$
    \State $T_2 \leftarrow \frac{(1 - \lambda) N}{\mu_{22} \cdot I_2 \cdot \beta_2}$
        
    \State \textbf{Step 3: Compute Data Transmission Latency}
    \State $T_3 \leftarrow \frac{\ N_e}{\beta_s}$
        
    \State \textbf{Step 4: Calculate Total System Latency}
    \State $T_{\text{total}}(\lambda) \leftarrow T_1 + T_2 + T_3$
    
    \State \textbf{Step 5: Update Partition Point}
    \State $\lambda \leftarrow \lambda - \alpha \cdot \nabla_{\lambda} T_{\text{total}}(\lambda)$
    
    \State \textbf{Step 6: Adjust Computational Intensity If Needed}
    \If{$I_1$ does not fully utilize Device 1's capacity}
        \State Adjust $\lambda$ or $I_0$ to maximize $I_1$ to achieve higher performance.
    \EndIf

\EndFor
\State \Return Optimized partition point $\lambda^*$, total system latency $T_{\text{total}}^*$
\end{algorithmic}
\end{algorithm}

\textit{System Latency Analysis}. System latency analysis involves the computation latency of the model and the data transmission latency between devices. We partition the model strategically to maximize the computational performance of the hardware devices, thereby reducing inference time, as shown in Algorithm~\ref{alg:latency_optimization}. The detailed process is as follows. The computation latency \( T_1 \) of sub-model 1 depends on its task load and the computational performance of hardware 1, expressed as:

\begin{equation}
T_1 = \frac{\lambda \cdot N}{\mu_{1} \cdot I_1 \cdot \beta_1},
\end{equation}
where \( \lambda \) represents the partition point of the complete model, \( N \) denotes the total computational load of the model, and \( \mu_1 \) represents the percentage of peak computational performance utilized by hardware 1. The computation latency \( T_2 \) of sub-model 2 depends on its task load and the computational performance of hardware 2, expressed as:

\begin{equation}
T_2 = \frac{(1 - \lambda) \cdot N}{\mu_{2} \cdot I_2 \cdot \beta_2}.
\end{equation}

The data transmission latency \( T_3 \) between devices is determined by the volume of data transmitted, specifically the \textit{embeddings} \( N_e \) output by the neural network, and the system transmission bandwidth \( \beta_s \). The total system latency \( T_{\text{total}}(\lambda) \) is the cumulative sum of the computation latency from the two sub-models and the data transmission latency between devices:

\begin{equation}
\begin{aligned}
T_{\text{total}}(\lambda) &= T_1 + T_2 + T_3 \\
&= \frac{\lambda N}{\mu_{1} \cdot I_1 \cdot \beta_1} + \frac{(1 - \lambda) N}{\mu_{2} \cdot I_2 \cdot \beta_2} + \frac{\ N_e}{\beta_s}.
\end{aligned}
\end{equation}

\textit{The Re-parameterization Model Structure}. We introduce the re-parameterization strategy into model construction, designing a dynamic network structure based on the RepVGG architecture that allows for structural variability during the inference phase.\footnote{https://github.com/word-ky/Co-design-Sys-NN} Unlike neural architecture search (NAS), this strategy enables the network to adaptively adjust its structure during inference to fit the hardware device. It offers greater flexibility in aligning the model with computational resources. The network can be fully re-parameterized, such as traditional RepVGG\cite{ding2021repvgg}, which fuses multiple channels into a single 3*3 convolutional layer, thereby reducing the computational burden on resource-constrained devices. In addition to RepVGG, we apply this strategy to other commonly used neural networks for edge deployment, such as ResNet-50 and GoogLeNet (see TABLE \ref{table:rep-model}). We use computational intensity to measure performance changes, and due to the greater variety of convolutional kernels in GoogLeNet, the computational intensity of re-parameterized Rep-GoogLeNet changes significantly. 

\begin{table}[h!]
\centering
\caption{Computational intensity variations in re-parameterized models. The re-parameterization strategy is applied to commonly used neural networks for edge deployment, with the improved models denoted as \textbf{Rep-X}. '\textbf{Original Intensity}' refers to the computational intensity of the original model, while '\textbf{Fully Reparameterization Intensity}' indicates the intensity after all channels are merged into a single fused channel. These metrics are used to quantify the performance variation in \textbf{Rep-X} models.}
\setlength{\tabcolsep}{0.15cm}
\begin{tabular}{l|c|>{\columncolor[gray]{0.9}}c}
\toprule
\textbf{Model}       & \makecell{\textbf{Original Intensity} \\ \textbf{(FLOPs/Byte)}} & \makecell{\textbf{Fully Reparameterization } \\ \textbf{Intensity (FLOPs/Byte)}} \\ \midrule
\textbf{RepVGG}     & 19.81                                   & 14.29                                    \\ \midrule
\textbf{Rep-ResNet-50}  & 38.78                               & 34.65                                    \\ \midrule
\textbf{Rep-ResNet-101} & 44.44                               & 38.73                                    \\ \midrule
\textbf{Rep-ResNet-152} & 49.13                               & 42.56                                    \\ \midrule
\textbf{Rep-MobileNetV1} & 11.47                              & 10.38                                    \\ \midrule
\textbf{Rep-ShuffleNet V1} & 9.62                              & 8.18                                    \\ \midrule
\textbf{Rep-EfficientNet} & 10.48                              & 8.52                                    \\ \midrule
\textbf{Rep-GoogLeNet}  & 32.43                               & 24.89                                    \\ \hline
\end{tabular}
\label{table:rep-model}
\end{table}

In contrast, high computational complexity models can improve throughput by maximizing the peak computational power of the device\cite{williams2009roofline}. To address this conflict, we propose a dynamic neural network fusion strategy. Unlike traditional full re-parameterization methods, this approach performs adaptive network fusion based on the device's capabilities. This allows the model to maximize performance without incurring excessive computational overhead. Below we provide a detailed explanation of this strategy. Following the RepVGG framework, the model block with a three-channel structure, consisting of a shortcut, a 3×3 convolution channel, and a 1×1 convolution channel, is designed, as illustrated in Fig. \ref{fig:dynamicmodelstructure}. Therefore, for each sub-model, the following four network fusion strategies \(\theta \) are considered:

\begin{itemize}
    \item Retain only the 3×3 convolution, denoted as \(S_3\).
    \item Retain both the 3×3 convolution and the shortcut, denoted as \(S_3 + S_s\).
    \item Retain the 3×3 convolution and the 1×1 convolution, denoted as \(S_3 + S_{1}\).
    \item Retain the 3×3 convolution, the shortcut, and the 1×1 convolution, denoted as \(S_3 + S_s + S_{1}\).
\end{itemize}

\textit{The Analysis of Dymanic Model Structure}. First, we analyze the impact of the dynamic model structure on system latency. When different network fusion strategies are applied to each sub-model, the computational load varies significantly, which greatly affects the inference speed. Specifically, the inference latency increases as the number of channels grows. Therefore, the model's network structure should be considered when calculating system latency. We improve the total system latency \( T_{\text{total}}(\lambda) \) as follows:
\begin{equation}
T_{\text{total}}(\lambda, \theta_1, \theta_2) = \frac{\lambda N(\theta_1)}{\mu_1 \cdot I_1 \cdot \beta_1} + \frac{(1 - \lambda) N(\theta_2)}{\mu_2 \cdot I_2 \cdot \beta_2} + \frac{N_e}{\beta_s}
\end{equation}
Where the \( \theta_1 \) and \( \theta_2 \) denote the sub-model fusion strategies, and \( N(\theta_1) \) and \( N(\theta_2) \) represent the computational loads of the sub-models.

Moreover, we analyze the impact of the dynamic model structure on accuracy, summarizing the computation process of accuracy loss for sub-models in Algorithm~\ref{alg:accuracy_loss}.

\begin{algorithm}
\caption{Computation of Accuracy Loss for Sub-models}
\label{alg:accuracy_loss}
\begin{algorithmic}[1]
\Statex \textbf{Input:} Partition point $\lambda$, sub-models fusion strategies $\theta_1$, $\theta_2$
\Statex \textbf{Output:} Total accuracy loss $\Delta A_{\text{total}}(\lambda, \theta_1, \theta_2)$

\State \textbf{Step 1: Calculate Accuracy Loss for Sub-model 1}
\State $\Delta A_1(\theta_1) \leftarrow$ Calculate accuracy loss using the network fusion strategy $\theta_1$ for sub-model 1

\State \textbf{Step 2: Calculate Accuracy Loss for Sub-model 2}
\State $\Delta A_2(\theta_2) \leftarrow$ Calculate accuracy loss using the network fusion strategy $\theta_2$ for sub-model 2

\State \textbf{Step 3: Calculate Total Accuracy Loss}
\State $\Delta A_{\text{total}}(\lambda, \theta_1, \theta_2) \leftarrow \lambda \cdot \Delta A_1(\theta_1) + (1 - \lambda) \cdot \Delta A_2(\theta_2)$

\State \Return Total accuracy loss $\Delta A_{\text{total}}(\lambda, \theta_1, \theta_2)$
\end{algorithmic}
\end{algorithm}

According to the reparameterization theory, the performance should not significantly degrade after network fusion. However, in practice, removing the 1×1 convolution channel has been shown to negatively affect the model's ability to perceive local small-sized features. Similarly, the shortcut channel impacts gradient computation in deeper networks, which is also reflected in the model’s accuracy. We use the \textit{Sign} function to represent the negative impact of different fusion strategies on performance, expressed as $\Delta A_n^m$. For different sub-models, the impact on accuracy is expressed accordingly,
\begin{equation}
\begin{aligned}
&\Delta A_n(\theta_n) = \sum_{m=1}^{4} \text{\textit{Sign}}(S_m) \cdot \Delta A_n^m\\
&= \text{\textit{Sign}}(S_3) \cdot \Delta A_n^1 + \text{\textit{Sign}}(S_3 + S_s) \cdot \Delta A_n^2 + \text{\textit{Sign}}(S_3 + S_{1x1}) \\
&\cdot \Delta A_n^3 + \text{\textit{Sign}}(S_3 + S_s + S_{1x1}) \cdot \Delta A_n^4,
\end{aligned}
\end{equation}
where $n \in \{1, 2\}$ denotes varrying sub-models and $m \in \{1, 2, 3, 4\}$ denotes above four fusion strategies. The total accuracy loss for the model is expressed as the weighted sum of the accuracy losses of sub-model 1 and sub-model 2:

\begin{equation}
\Delta A_{\text{total}}(\lambda, \theta_1, \theta_2) = \lambda \cdot \Delta A_1(\theta_1) + (1 - \lambda) \cdot \Delta A_2(\theta_2),
\end{equation}
where the \( \lambda \) represents the partition point of the whole model.

\begin{algorithm}
\caption{Joint Optimization for Latency and Accuracy}
\label{alg:joint_optimization_grid_search}
\begin{algorithmic}[1]
\Statex \textbf{Input:} Set of possible partition points $\Lambda$, set of possible fusion strategies $\Theta_1$, $\Theta_2$, Lagrangian multiplier $\lambda_1$
\Statex \textbf{Output:} Optimized partition point $\lambda^*$, optimized network fusion strategies $\theta_1^*$, $\theta_2^*$

\State Initialize the best latency-accuracy trade-off score $L^* \leftarrow \infty$, best partition point $\lambda^*$, best fusion strategies $\theta_1^*$, $\theta_2^*$

\For{each $\lambda \in \Lambda$}
    \For{each $\theta_1 \in \Theta_1$}
        \For{each $\theta_2 \in \Theta_2$}

            \State \textbf{Step 1: Calculate System Latency}
            \State $T_{\text{total}}(\lambda, \theta_1, \theta_2) \leftarrow$ Compute system latency based on the partition point and sub-model structure
            
            \State \textbf{Step 2: Compute Total Accuracy Loss}
            \State $\Delta A_{\text{total}}(\lambda, \theta_1, \theta_2) \leftarrow$ Use the result from Algorithm~\ref{alg:accuracy_loss}
            
            \State \textbf{Step 3: Calculate the Joint Optimization Function}
            \State $L(\lambda, \theta_1, \theta_2, \lambda_1) \leftarrow T_{\text{total}}(\lambda, \theta_1, \theta_2) + \lambda_1 \cdot \Delta A_{\text{total}}(\lambda, \theta_1, \theta_2)$
            
            \State \textbf{Step 4: Check for Optimal Solution}
            \If{$L(\lambda, \theta_1, \theta_2, \lambda_1) < L^*$}
                \State Update the best score $L^* \leftarrow L(\lambda, \theta_1, \theta_2, \lambda_1)$
                \State Update the optimal values $\lambda^* \leftarrow \lambda$, $\theta_1^* \leftarrow \theta_1$, $\theta_2^* \leftarrow \theta_2$
            \EndIf

        \EndFor
    \EndFor
\EndFor

\State \Return Optimized partition point $\lambda^*$, optimized fusion strategies $\theta_1^*$, $\theta_2^*$
\end{algorithmic}
\end{algorithm}

\textit{The multi-objective joint optimization framework.} The multi-objective joint optimization framework is proposed to achieve the co-optimization of latency and accuracy. Specifically, by partitioning the model and applying the optimal network fusion strategies to different sub-models, the system's latency and accuracy loss can be minimized while meeting computational intensity requirements. The multi-objective co-optimization framework is defined as a \textit{Lagrangian joint optimization function \( L \)}, which combines total system latency and the negative impact on accuracy.

\begin{equation}
\min_{\lambda, \theta_1, \theta_2, \lambda_1} L(\lambda, \theta_1, \theta_2, \lambda_1) = T_{\text{total}}(\lambda, \theta_1, \theta_2) + \lambda_1 \cdot \Delta A_{\text{total}}(\lambda, \theta_1, \theta_2),
\end{equation}
where \( T_{\text{total}}(\lambda) \) is the total system latency, \( \Delta A_{\text{total}}(\lambda, \theta_1, \theta_2) \) is the total accuracy loss, and \( \lambda_1 \) is a Lagrangian multiplier that controls the trade-off between latency and accuracy loss. By adjusting the partition point \( \lambda \) and the network fusion strategies \( \theta_1 \), \( \theta_2 \), we can find the optimal solution that ensures computational efficiency while minimizing accuracy loss. The optimization processing as shown in the Algorithm~\ref{alg:joint_optimization_grid_search}, for the optimization approach, since both the partition point and fusion strategies are limited, that is, \( \theta_1 \), \( \theta_2 \), and \( \lambda \) have fixed ranges of values (\( \lambda \in \Lambda \), \( \theta_1 \in \Theta_1 \), and \( \theta_2 \in \Theta_2 \)), we use a simple grid search as an illustrative example. In practice, optimization goals can also be achieved through methods such as parameter search strategies based on gradient descent\cite{xue2022self, arab2015adaptive}.

\section{Convergence Analysis of The Partitioned Model}
\label{sec:convergence}

In this section, we conduct a convergence analysis for the optimization of the base model at partitioned deployment. The optimized model will be input into the proposed co-design method, where it serves as the initial state for co-optimization. First, we analyze the mathematical consistency of parameter updates between the partitioned model and the original full model. Furthermore, we provide a detailed convergence proof for the partitioned model.

\textit{Consistency Analysis in terms of gradient computation and parameter updates.} By analyzing the consistency of gradient computation and parameter updates, we ensure that the split model remains mathematically equivalent to the original model. This analysis is crucial because if the splitting process introduces discrepancies, such as incorrect gradients or parameter updates, several issues may arise: 1) the model’s optimization process could deviate from the desired behavior; 2) the theoretical guarantees of convergence (e.g., achieving global or local optima) may no longer hold; and 3) the convergence proof derived for the non-split model would no longer apply to the split model, necessitating additional complex modifications.

The detailed analysis is as follows: in the split model, the gradients are computed independently by Device 1 and Device 2:

\begin{equation}
g_n = \nabla_{w_n} L(w_1, w_2), \quad n = 1, 2
\end{equation}
That is, Device 1 computes the partial derivative of the loss function \( L(w_1, w_2) \) with respect to the parameter \( w_1 \), and Device 2 computes the partial derivative with respect to the parameter \( w_2 \). Since the forward and backward propagation processes in the split model compute all gradient information completely, combining the gradients from Device 1 and Device 2 yields the overall gradient:
\begin{equation}
\nabla L(w) = \left[ g_1, g_2 \right],
\end{equation}
this is consistent with the overall gradient computation in the original model. Thus, the split model maintains consistency with the non-split model in terms of gradient computation.

Moreover, parameter updates are performed independently on Device 1 and Device 2:

\begin{equation}
w_n^{(k+1)} = w_n^{(k)} - \eta g_n, \quad n = 1, 2
\end{equation}
That is, Device 1 updates the parameter \( w_1 \) using the computed gradient \( g_1 \) and the learning rate \( \eta \), while Device 2 updates the parameter \( w_2 \) using the computed gradient \( g_2 \) and the learning rate \( \eta \). This update rule is also consistent with the overall parameter update rule in the original model. Therefore, the split model also maintains consistency with the full model in terms of parameter updates.

\textit{Convergence Proof for the Split Model.} Based on the above analysis of gradient computation and parameter update consistency, we ensure that the split model is mathematically equivalent to the original non-split model. Next, we will prove the convergence of the split model. 

\subsubsection{Parameter Update Rules.} In the split model, Devices 1 and 2 update parameters \( w_1 \) and \( w_2 \) independently, following the update rule:

\begin{equation}
w_n^{(k+1)} = w_n^{(k)} - \eta g_n = w_n^{(k)} - \eta \nabla_{w_n} L(w_1^{(k)}, w_2^{(k)}), \quad n = 1, 2
\end{equation}

We aim to derive the change in the loss function \( L(w_1, w_2) \) after each iteration, thereby proving the convergence of the split model.

\subsubsection{Loss Function Reduction Using Strong Convexity assumption}
Based on the mathematical consistency between the complete model and the split model, we employ the assumption of \textbf{strong convexity}:

\begin{equation}
L(w') \geq L(w) + \nabla L(w)^T (w' - w) + \frac{\mu}{2} \|w' - w\|^2 .
\end{equation}

For the split model, assuming \( L(w_1, w_2) \) is strongly convex with respect to \( w_1 \) and \( w_2 \), for any \( w_1^{(k+1)}, w_2^{(k+1)} \) and \( w_1^{(k)}, w_2^{(k)} \), we have Eq. ~\eqref{eq:convergence1} where \( \mu > 0 \) is the strong convexity constant, and this formula describes the decrease in the loss function during each iteration.

\begin{figure*}[!t]
\begin{align}
L(w_1^{(k+1)}, w_2^{(k+1)}) \geq &L(w_1^{(k)}, w_2^{(k)}) + \nabla_{w_1} L(w_1^{(k)}, w_2^{(k)})^T (w_1^{(k+1)} - w_1^{(k)}) \nonumber \\
            &+ \nabla_{w_2} L(w_1^{(k)}, w_2^{(k)})^T (w_2^{(k+1)} - w_2^{(k)}) \nonumber \\
            &+ \frac{\mu}{2} \left(\|w_1^{(k+1)} - w_1^{(k)}\|^2 + \|w_2^{(k+1)} - w_2^{(k)}\|^2\right).
\label{eq:convergence1}
\end{align}
\end{figure*}

\subsubsection{Substituting the Parameter Update Equations}

Substituting the gradient descent update rules \( w_1^{(k+1)} = w_1^{(k)} - \eta \nabla_{w_1} L(w_1^{(k)}, w_2^{(k)}) \) and \( w_2^{(k+1)} = w_2^{(k)} - \eta \nabla_{w_2} L(w_1^{(k)}, w_2^{(k)}) \) into the strong convexity expression, as shown in Eq. ~\eqref{eq:convergence2}, this expression demonstrates the change in the loss function after each iteration. To ensure that the loss function decreases after each iteration, we require that the right-hand side of this expression is negative.

\begin{figure*}[!t]
\begin{equation}
\begin{aligned}
\longrightarrow L(w_1^{(k+1)}, w_2^{(k+1)}) \geq &L(w_1^{(k)}, w_2^{(k)}) - \eta \|\nabla_{w_1} L(w_1^{(k)}, w_2^{(k)})\|^2 - \eta \|\nabla_{w_2} L(w_1^{(k)}, w_2^{(k)})\|^2 \\
&+ \frac{\mu}{2} {\eta^2} \left( \|\nabla_{w_1} L(w_1^{(k)}, w_2^{(k)})\|^2 + \|\nabla_{w_2} L(w_1^{(k)}, w_2^{(k)})\|^2 \right).
\end{aligned}
\label{eq:convergence2}
\end{equation}
\end{figure*}

\subsubsection{Ensuring Loss Function Reduction}

To ensure that the loss function decreases after each iteration, we require Eq. ~\eqref{eq:convergence3}. Factoring out the gradient norms, we get Eq. ~\eqref{eq:convergence4}. Furthermore, since the gradient norm squared \( \|\nabla_{w_1} L(w_1^{(k)}, w_2^{(k)})\|^2 + \|\nabla_{w_2} L(w_1^{(k)}, w_2^{(k)})\|^2 \) is non-negative, we can drop it, yielding the condition:

\begin{figure*}[!t]
\begin{equation} 
\longrightarrow -\eta \|\nabla_{w_1} L(w_1^{(k)}, w_2^{(k)})\|^2 - \eta \|\nabla_{w_2} L(w_1^{(k)}, w_2^{(k)})\|^2 + \frac{\mu}{2} \eta^2 \left(\|\nabla_{w_1} L(w_1^{(k)}, w_2^{(k)})\|^2 + \|\nabla_{w_2} L(w_1^{(k)}, w_2^{(k)})\|^2\right) < 0.
\label{eq:convergence3}
\end{equation}
\end{figure*}

\begin{figure*}[!t]
\begin{equation} 
\longrightarrow \left(-\eta + \frac{\mu}{2} \eta^2\right) \left(\|\nabla_{w_1} L(w_1^{(k)}, w_2^{(k)})\|^2 + \|\nabla_{w_2} L(w_1^{(k)}, w_2^{(k)})\|^2\right) < 0.
\label{eq:convergence4}
\end{equation}
\end{figure*}

\begin{equation}
-\eta + \frac{\mu}{2} \eta^2 < 0.
\end{equation}

\subsubsection{Deriving the Range of the Learning Rate}

To satisfy this inequality, we solve for the range of the learning rate \( \eta \). This inequality can be treated as a quadratic equation in \( \eta \):

\begin{equation}
\frac{\mu}{2} \eta^2 - \eta < 0.
\end{equation}

\begin{equation}
\longrightarrow \eta \left( \frac{\mu}{2} \eta - 1 \right) < 0.
\end{equation}

This implies that the learning rate \( \eta \) must satisfy:

\begin{equation}
0 < \eta < \frac{2}{\mu}.
\end{equation}

This is the lower bound on the learning rate that ensures the loss function decreases after each iteration.

\subsubsection{Determining the Upper Bound of the Learning Rate Using Lipschitz Continuity}

According to the assumption of \textbf{Lipschitz continuity}, the gradient of the loss function satisfies the following condition: there exists a constant \( L > 0 \) such that for all \( w \) and \( w' \):

\begin{equation}
\|\nabla L(w) - \nabla L(w')\| \leq L \|w - w'\|.
\end{equation}

This implies that the gradient variation is bounded. To ensure that the gradient change is not too large, and that the loss function decreases effectively after each iteration, the upper bound of the learning rate \( \eta \) should be:

\begin{equation}
0 < \eta \leq \frac{2}{L}.
\end{equation}

This upper bound controls the step size in each iteration, preventing instability or divergence in convergence. Under the above conditions, the loss function \( L(w_1, w_2) \) in the split model will converge to the optimal value after several iterations. After \( K \) iterations, the gap between the loss function and the optimal solution satisfies Eq. ~\eqref{eq:convergence5}, this indicates that the gap between the current solution and the optimal solution decreases by a fixed proportion after each iteration.

\begin{figure*}[!h]
\begin{equation} 
L(w_1^{(k)}, w_2^{(k)}) - L(w_1^*, w_2^*) \leq \left(1 - \frac{\eta \mu}{2} \right)^k \left( L(w_1^{(0)}, w_2^{(0)}) - L(w_1^*, w_2^*) \right).
\label{eq:convergence5}
\end{equation}
\end{figure*}

\section{Experiments \& Analysis}

\subsection{Experimental setting}

\textit{Devices}. We selected four commonly used resource-constrained embedded devices\cite{mittal2019survey}, namely NVIDIA Jetson Nano (Nano), NVIDIA Jetson TX1 (TX1),  NVIDIA Jetson TX2 (TX2) and NVIDIA Jetson Xavier NX (NX) as experimental devices. The system transmission bandwidth is set to 100 Mbps. In each experiment, two of these devices were selected to serve as the computational terminal (Device 1) and the edge server (Device 2) in the IoVT system, arranged in ascending order of computational intensity. Moreover, we use TensorRT, an NVIDIA deep learning acceleration tool, to enhance the inference speed.

\textit{Datasets and models}. The experimental datasets include MNIST\cite{deng2012mnist}, CIFAR-100\cite{sharma2018analysis}, and ImageNet\cite{deng2009imagenet}. We utilize the default training and test set splits of these datasets, respectively. During the testing phase, we simulated real-world random event intervals by driving edge inference tasks with user requests following an exponential distribution\cite{guo2009analyzing}. This approach allows for a more realistic simulation of system behavior under uncertain conditions, assessing how edge devices handle fluctuating request frequencies and enabling us to evaluate system performance and reliability. To verify the generalizability of the proposed method across different models, we selected RepVGG, Rep-ResNet, and Rep-GoogLeNet as the base models, as listed in TABLE \ref{table:rep-model}. The following TABLE \ref{Table:computingintensityofrepmodel} compares the computational intensity of the models and devices. 

\begin{table}[h!]
\centering
\setlength{\tabcolsep}{0.15cm}
\caption{The comparison of computing intensity between models and devices. If the model's computing intensity exceeds that of the devices, we call it "\textbf{Computing Constraint (CC)}," indicating that the model is limited by the device's maximum computational power. Conversely, if the device's computing intensity is higher, the model is constrained by memory, referred to as "\textbf{Memory Constraint (MC)}."} 
\begin{tabular}{lccccc}
\toprule
\textbf{Model}     & \makecell{\textbf{Computing Intensity} \\ \textbf{(FLOPs/Byte)}} & \textbf{Nano} & \textbf{TX1} & \textbf{TX2} & \textbf{NX}  \\ \midrule
\textbf{RepVGG}     & 19.81                                   & CC    & CC    & MC    & MC   \\ \midrule
\textbf{Rep-ResNet-50}  & 38.78                                  & CC   & MC   & MC   & MC   \\ \midrule
\textbf{Rep-GoogLeNet}  & 32.43                                   & CC    & MC    & MC    & MC   \\ \bottomrule
\label{Table:computingintensityofrepmodel}
\end{tabular}
\end{table}

\textit{Evaluate metrics and goals}. We employed three key performance metrics to evaluate the real-time performance and accuracy of the IoVT system. \textbf{Throughput}: This refers to the number of requests the system can process within a unit of time. A higher throughput indicates that the system can handle more tasks in a shorter period, demonstrating good real-time performance. \textbf{Response time} is another indicator of real-time performance, complementing throughput by evaluating the method's real-time capability from a temporal perspective. \textbf{Accuracy}: This metric evaluates the correctness of classification tasks, specifically the proportion of correctly classified instances in image classification. \textbf{Acc-X}: This metric represents the accuracy for specific categories and is measured using \textit{mAP@IoU}=0.5. It reflects the mean Average Precision (mAP) at an Intersection over Union (IoU) threshold of 0.5, which is used to evaluate the overall performance of the model in detection tasks.

We evaluate our method by addressing the following research questions:

\begin{itemize}
    \item \textbf{RQ1 (Real-time):} Can our method improve throughput or reduce response time to achieve real-time visual inference on edge devices with limited resources?
    \item \textbf{RQ2 (Accuracy):} Can our method achieve high classification accuracy and demonstrate excellent detection accuracy for small objects in object detection tasks?
    \item \textbf{RQ3 (Generalizability):} Can our method deploy the same base model across different embedded devices while maintaining high throughput and accuracy? Can it perform consistently well on various base models and datasets?
    \item \textbf{RQ4 (Practical Application):} Can our method assist IoVT systems in performing real-time and precise edge inference?
\end{itemize}

\textit{Experiment Process}. The experimental process can be divided into "Comparative experiment", "Ablation experiment" and "Case Analysis", and we use the results to answer all four RQs.

\textbf{Comparative experiment.} We utilize common image classification tasks to conduct comparative experiments, evaluating the performance of our method against six state-of-the-art (SOTA) algorithms on the MNIST, CIFAR-100, and ImageNet datasets. The experimental devices are configured as follows: For the computing terminal (Device 1), we utilize the NVIDIA Jetson Nano, which has a peak computational power of 472 GFLOPS, a memory bandwidth of 25.6 GB/s, and is equipped with a 128-core NVIDIA Maxwell™ architecture GPU. For the edge server (Device 2), which handles both training and inference tasks, we employ the more powerful NVIDIA Jetson Xavier NX. This device has a peak computational power of 21 TOPS, a memory bandwidth of 51.2 GB/s, and features a 384-core NVIDIA Volta™ architecture GPU. RepVGG is selected as the base model, and we apply TensorRT acceleration strategies to conduct inference for each method. If the comparison algorithms do not involve partitioning strategies, inference is performed entirely on the edge server. Additionally, the hyperparameters for all methods are fine-tuned to ensure optimal performance. As outlined in Section \ref{sec:convergence}, RepVGG is partitioned and deployed across Device 1 and Device 2 for training, which serves as the initial state for subsequent co-optimization. The other training setup follows the configuration described in \cite{sidd}. 

For \textbf{RQ1}, we conduct real-time performance experiments, using throughput and response time as metrics to assess the inference speed and latency, testing whether our method can achieve real-time edge inference (see Fig. \ref{fig:comparison-realtime}). For \textbf{RQ2}, we perform accuracy experiments on the three datasets (see Fig. \ref{fig:comparison-acc}) to evaluate whether our method can achieve high classification accuracy.

\textbf{Ablation Experiments.} We conduct ablation experiments to evaluate the performance of the strategies proposed in our method, using the three datasets and three re-parameterized models, as shown in TABLE \ref{Table:computingintensityofrepmodel}. Additionally, we further assess the advantages of our method, such as its generalizability.

For \textbf{RQ1} and \textbf{RQ2}, in the ablation experiments, we set conditions such as "w/o roofline strategy," "w/o dynamic model structure," and "w/o multi-objective function" to evaluate the impact of the proposed strategies on throughput and accuracy (see TABLE \ref{tab:abl-all}). Furthermore, we perform a visual ablation experiment specifically for the "dynamic model structure," demonstrating the differences in throughput caused by different partition points to further verify the effectiveness of the dynamic model structure. For \textbf{RQ3}, we use two typical distributed deployment methods, SIDD and RoofSplit, as comparison algorithms to evaluate the performance of our method with different base models on various datasets (see TABLE \ref{tab:abl-all}). Additionally, we assess the performance stability of our method using the same base model across different devices, including the previously mentioned Nano and NX, as well as the NVIDIA Jetson TX1, which has a peak computational power of 1 TFLOPS, a memory bandwidth of 25.6 GB/s, and a 256-core NVIDIA Maxwell™ architecture GPU; and the NVIDIA Jetson TX2, which has a peak computational power of 1.33 TFLOPS, a memory bandwidth of 59.7 GB/s, and a 256-core NVIDIA Pascal™ architecture GPU. The results of these experiments validate the generalizability of our method (see Fig. \ref{fig:abl-generdevice}).

\textbf{Case analysis.} For \textbf{RQ4}, we conduct a coal mine helmet object detection simulation experiment to assess the practical application value of our method. We use the YOLO framework with a re-parameterized backbone as the base model and contribute a new coal mine object detection dataset. The experimental setup includes TX2 as Device 1 and NVIDIA RTX 4090 as Device 2, with a network environment simulating an IoVT system using an uplink bandwidth of 107.4 Mbps. We evaluate detection accuracy and throughput using SIDD, RoofSplit, and our method with different computing approaches, such as cloud computing (see TABLE \ref{tab:simula-comparison}). Importantly, for \textbf{RQ2}, we further visualize the advantages of small object detection in Fig.\ref{fig:simul-comparisonvisual} and conduct "perceptual performance experiments" and "feature extraction performance experiments" to analyze our method's ability to extract detailed features and its performance in detecting small objects.


\subsection{Comparative experiment}

We conduct comparative experiments to evaluate the performance differences between our method and six state-of-the-art (SOTA) algorithms—SIDD\cite{sidd}, RoofSplit\cite{roofsplit}, MPMS\cite{mpms}, SciNet\cite{scinet}, SECDA\cite{secda}, and ADIA\cite{adia}—on the MNIST, CIFAR-100, and ImageNet datasets. 

\begin{figure*} [t!]
	\centering
           
        \subfloat[MNIST]{
	\includegraphics[width=5.8cm,height=4.4cm]{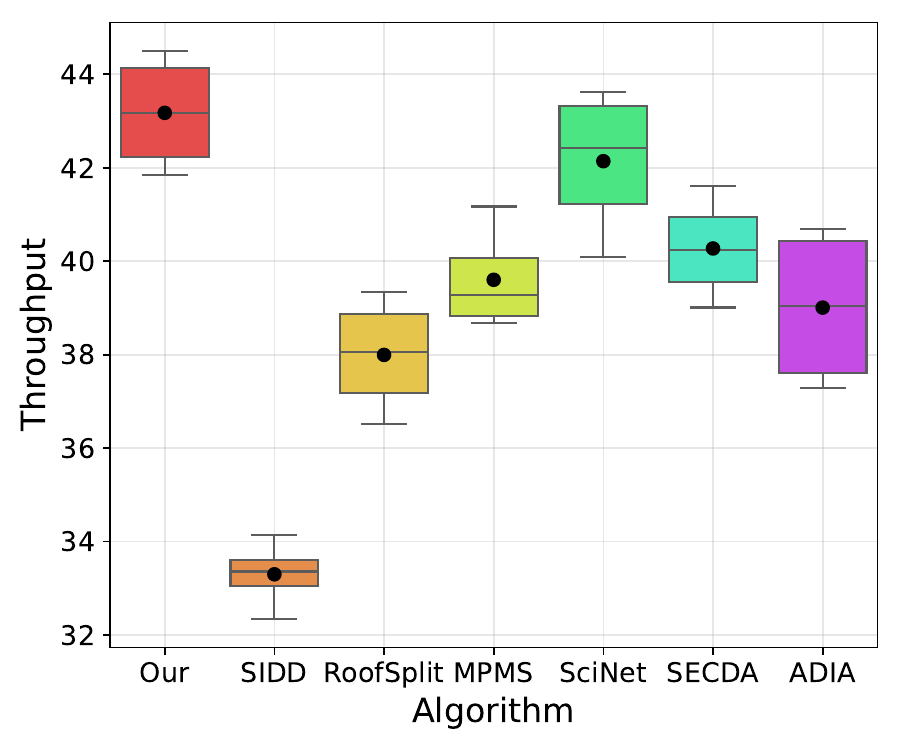}}
        \subfloat[CIFAR-100]{
	\includegraphics[width=5.8cm,height=4.4cm]{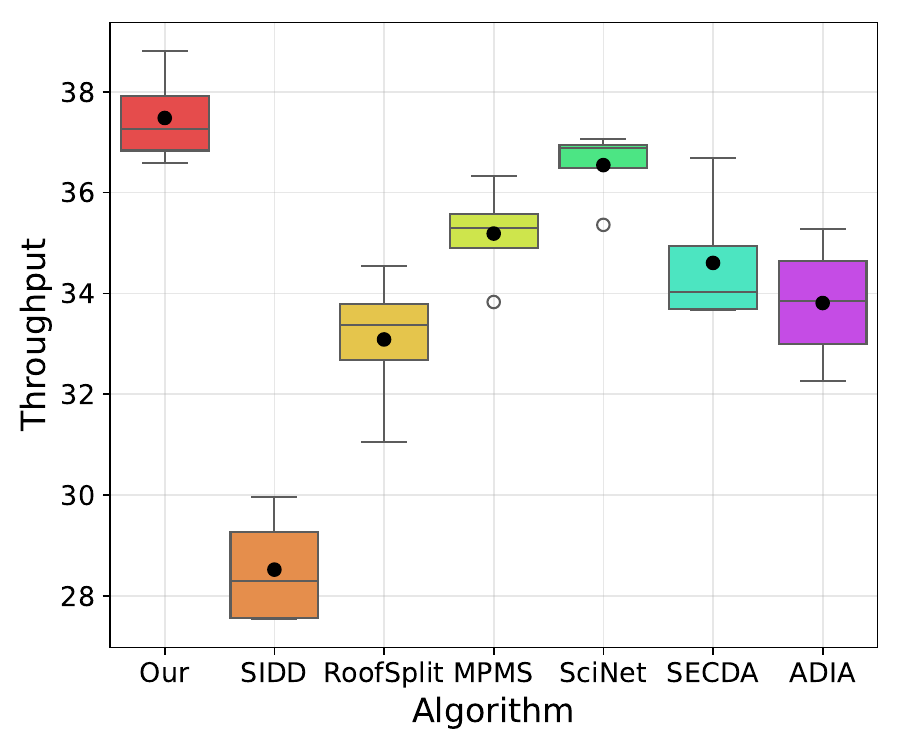}}
         \subfloat[ImageNet]{
	\includegraphics[width=5.8cm,height=4.4cm]{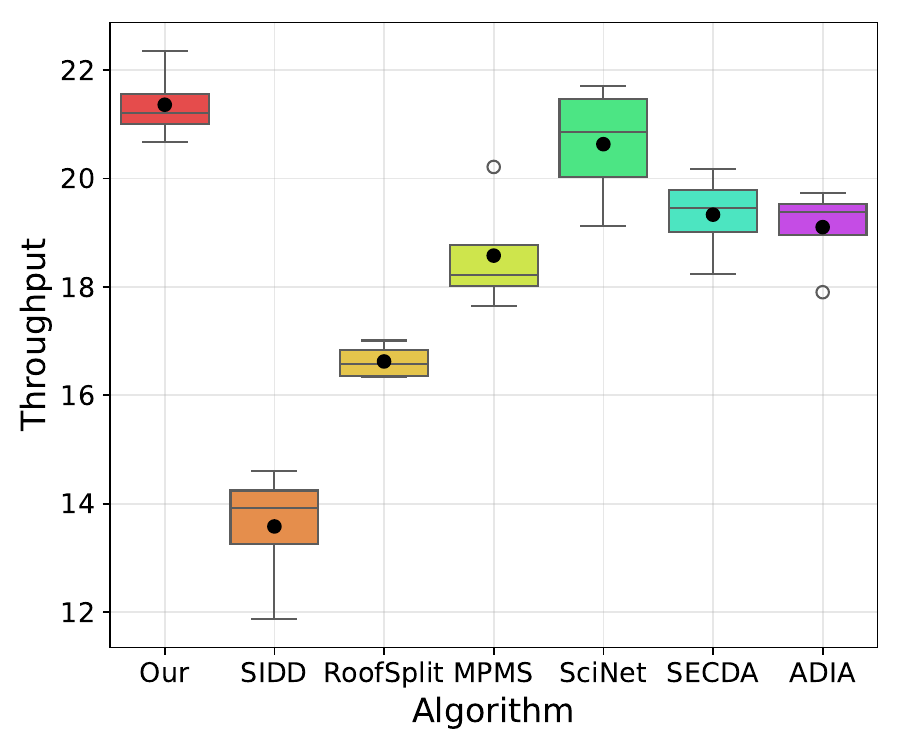}}\\
        \centering{Throughput}\\
        
            \subfloat[MNIST]{
	\includegraphics[width=5.8cm,height=4.4cm]{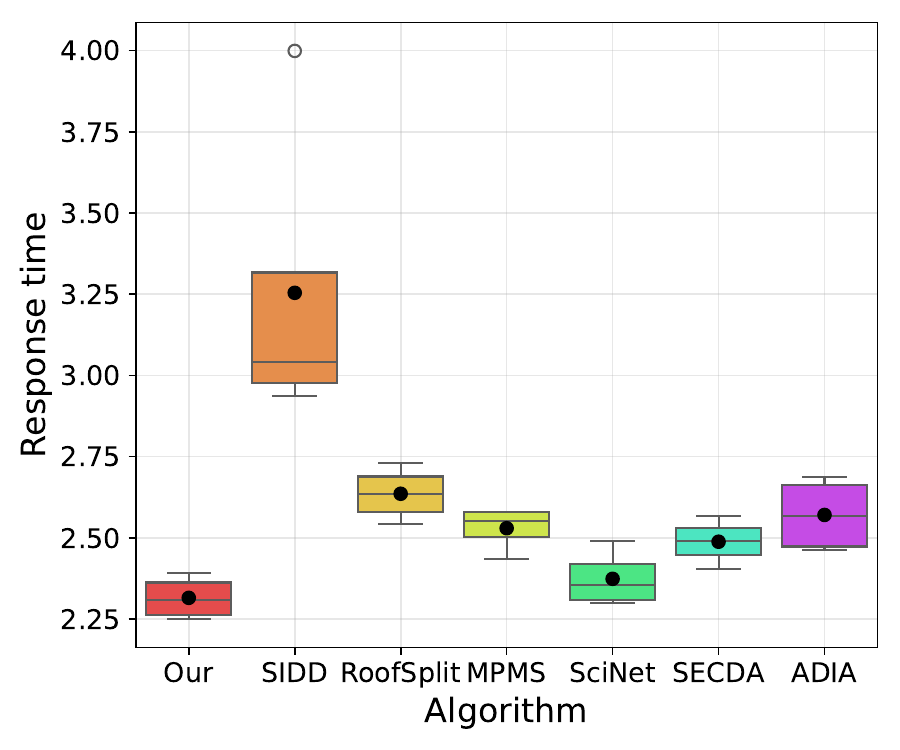}}
         \subfloat[CIFAR-100]{
		\includegraphics[width=5.8cm,height=4.4cm]{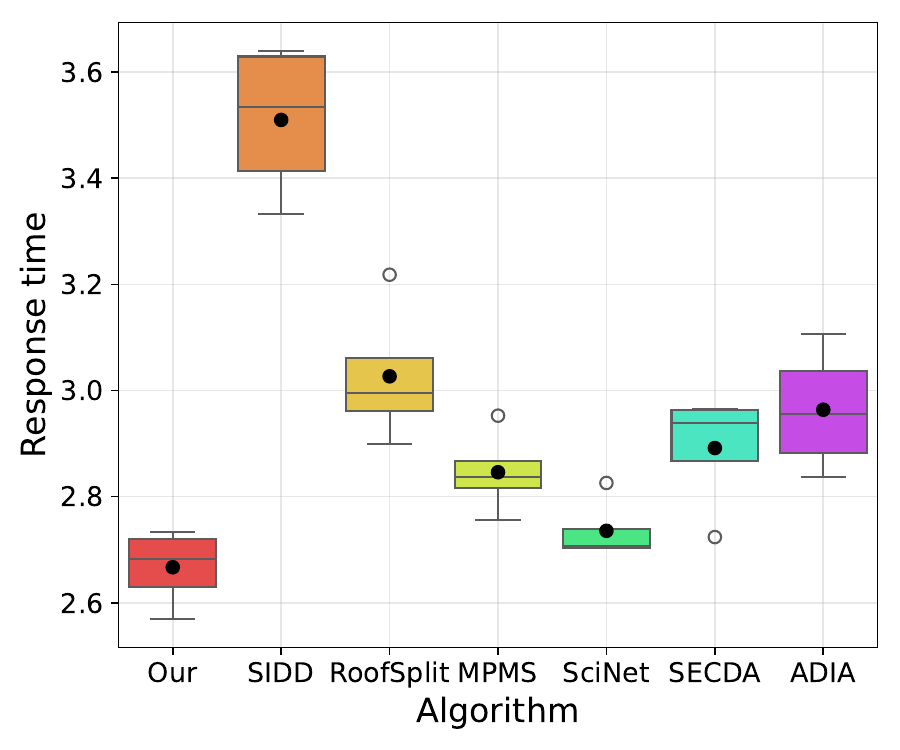}}
         \subfloat[ImageNet]{
		\includegraphics[width=5.8cm,height=4.4cm]{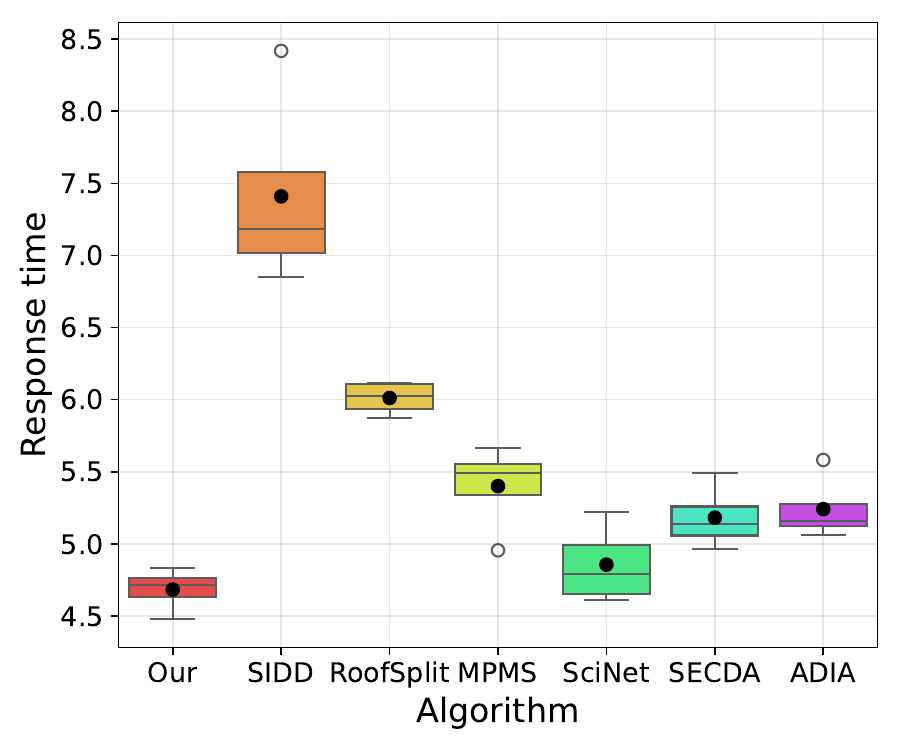}}\\
       
  \centering{Response time}\\
        
  \caption{The Comparison of real-time performance. "Our" denotes our method. In the boxplots presented in this experiment, black dots indicate the mean values, and hollow circles denote outliers. The response time per image for each method is expressed using the notation “$*10^2$”. Compared to the baseline algorithms, our approach demonstrates the highest throughput and faster response times, consistently achieving real-time inference performance, even on large-scale ImageNet images.
}
\label{fig:comparison-realtime}
\end{figure*}

\textit{Real-time performance analysis}. A real-time performance analysis was conducted to compare the proposed method with six state-of-the-art (SOTA) approaches across three datasets, evaluating key metrics including throughput and response time. The experimental results are detailed in Fig. \ref{fig:comparison-realtime}. Compared to the baseline algorithms, our method exhibits outstanding performance in both metrics across all three datasets. Specifically, for throughput, on the MNIST dataset, relative to RoofSplit (37.48 images/second), our method's throughput is higher by 15.4\%. Additionally, our method outperforms the second-best MPMS (39.50 images/second) with a 9.5\% increase in throughput. On CIFAR-100, our method surpasses MPMS (34.62 images/second) by 10.2\%. On the larger-scale ImageNet dataset, our method attains a throughput of 21.11 images/second, meeting real-time standards and outperforming SIDD (11.87 images/second) with a 77.9\% improvement. For response time, on MNIST, compared to RoofSplit (0.0612 seconds), our method achieves a 22.5\% faster response time. Moreover, our method outperforms MPMS (0.0546 seconds) by 13.2\%. On ImageNet, compared to RoofSplit (0.0517 seconds), our method is 10.3\% faster. Finally, our method improves upon MPMS (0.0489 seconds) with a 5.1\% faster response time. These results demonstrate that our method can achieve real-time edge inference.

\begin{figure}[t]
    \centering
    \includegraphics[width=\linewidth]{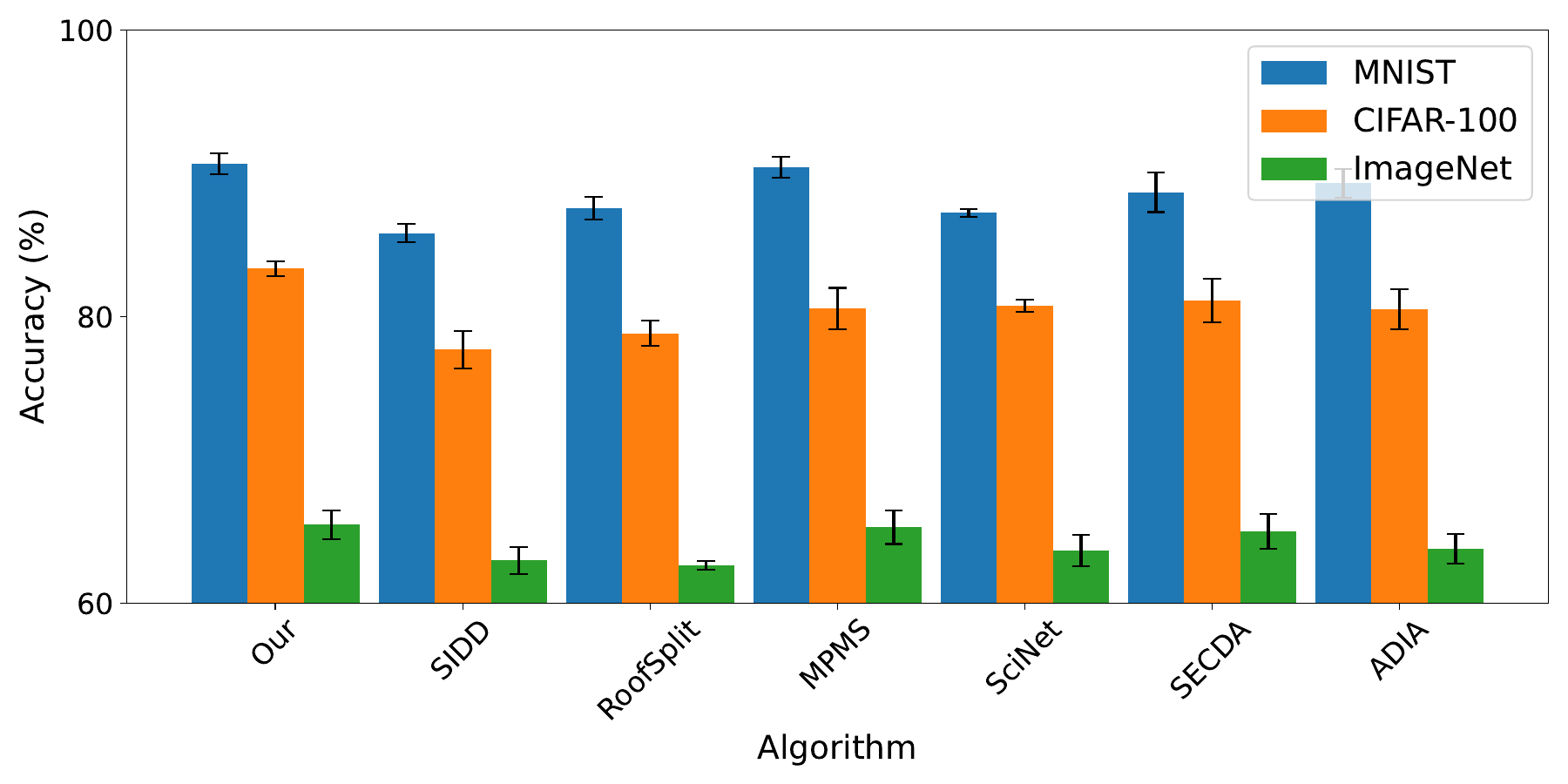}
    \caption{The comparison of accuracy. "Our" denotes our method. Our algorithm demonstrates a significant improvement in accuracy across all datasets compared to the baseline algorithms, with a particularly large lead on the CIFAR-100 dataset.
}
\label{fig:comparison-acc}
\end{figure}

\textit{Accuracy results analysis}. The experimental results are shown in Fig. \ref{fig:comparison-acc}. Compared to the baseline algorithms, our method maintains an accuracy advantage on the MNIST, CIFAR-100, and ImageNet datasets. Specifically, on the MNIST dataset, our method improves accuracy by 1.98 compared to SECDA (88.68\%). On the CIFAR-100 dataset, our method achieves an accuracy of 83.34\%, outperforming the second-best SECDA (81.64) by 1.7, and surpassing SciNet (80.75\%) by 2.59. On the ImageNet dataset, our method still reaches an accuracy of 65.47\%. These results demonstrate that our method can achieve outstanding classification accuracy at the edge side.

\subsection{Ablation Experiments}
We perform detailed ablation experiments on the proposed improvements to demonstrate the effectiveness of these methods. Additionally, we thoroughly evaluate the generalizability advantages of our method.

\begin{table*}[htbp]
\centering
\setlength{\tabcolsep}{0.12cm}
\caption{Ablation Experimental Results. In the experiments, "\textbf{w/o}" denotes the removal of a specific improvement from our method. "\textbf{w/o roofline strategy}" refers to using the base model in its initial partition state, where the entire model is deployed on the edge server (Device 2, NVIDIA Jetson Xavier NX). "\textbf{w/o dynamic model structure}" indicates that the sub-models deployed on different devices are in an initial random structure, rather than the optimized optimal structure. "\textbf{w/o multi-objective function}" signifies that both the model partition state and model structure remain in their initial states. Additionally, we use "\textbf{Average}" to represent the overall performance of the method. The experimental results show that our method is applicable across different base models, and the three main improvements contribute significantly to performance enhancement.}

\begin{tabular}{c|c|c|cc|cc|cc|c}
\hline
\multirow{2}{*}{{\textbf{\makecell{Experimental\\ Category}}}} & \multirow{2}{*}{{\textbf{Method}}} & \multirow{2}{*}{{\textbf{Model}}}& \multicolumn{2}{c|}{\textbf{MNIST}} & \multicolumn{2}{c|}{\textbf{CIFAR-100}} & \multicolumn{2}{c|}{\textbf{ImageNet}} & \multirow{2}{*}{\textbf{Average}} \\ 
       &       &       & \textbf{Throughput} & \textbf{Accuracy} & \textbf{Throughput} & \textbf{Accuracy} & \textbf{Throughput} & \textbf{Accuracy} &  \\ \hline
\multirow{9}{*}{\textbf{\makecell{Generalizability\\ Evaluation}}} &   & RepVGG         & 32.13 & 86.17 & 28.10 & 77.82 & 13.77 & 63.19 & 50.20 \\
    & {SIDD\cite{sidd}}  & Rep-Resnet     & 25.81 & 87.63 & 20.77 & 79.15 & 10.86 & 64.22 & 48.07 \\
    &       & Rep-GoogLeNet  & 22.15 & 89.17 & 15.93 & 82.17 & 8.22  & 65.06 & 47.12 \\ \cline{2-10}
    &   & RepVGG      & 36.93 & 88.25 & 33.18 & 77.62 & 16.43 & 63.28 & 52.62 \\
    & {RoofSplit\cite{roofsplit}}  & Rep-Resnet     & 34.26 & 90.28 & 29.84 & 81.49 & 15.37 & 64.87 & 52.69 \\
    &       & Rep-GoogLeNet  & 24.89 & 93.51 & 20.91 & 84.21 & 11.88 & 65.29 & 50.12 \\ \cline{2-10}
    &   & RepVGG     & 43.17 & 92.47 & 37.89 & 83.32 & 22.64 & 66.81 & 57.72 \\
    & {Our method}   & Rep-Resnet     & 35.82 & 94.79 & 32.15 & 84.70 & 20.58 & 67.72 & 55.96 \\
    &       & Rep-GoogLeNet  & 29.83 & 97.28 & 25.96 & 86.36 & 17.93 & 69.15 & 54.42 \\ \hline
\multirow{9}{*}{\textbf{\makecell{Improvements\\ Evaluation}}} &   & RepVGG & 41.72 & 89.04 & 35.72 & 81.15 & 19.36 & 65.77 & 55.46 \\
    & {w/o roofline strategy}   & Rep-Resnet     & 33.29 & 91.85 & 29.91 & 81.59 & 18.04 & 66.26 & 53.49 \\
    &       & Rep-GoogLeNet  & 28.97 & 93.73 & 25.01 & 84.33 & 16.83 & 68.59 & 52.91 \\ \cline{2-10}
    &   & RepVGG & 39.55 & 89.52 & 33.15 & 79.67 & 18.53 & 63.61 & 54.01 \\
    & {w/o dynamic model structure}   & Rep-Resnet     & 30.36 & 89.04 & 27.74 & 78.92 & 17.93 & 64.48 & 51.41 \\
    &       & Rep-GoogLeNet  & 22.17 & 90.37 & 19.48 & 82.56 & 12.64 & 65.76 & 48.83 \\ \cline{2-10}
    &   & RepVGG & 37.24 & 88.13 & 32.06 & 78.24 & 16.04 & 62.35 & 52.34 \\
    & {w/o multi-objective function}   & Rep-Resnet     & 28.95 & 88.49 & 25.64 & 77.19 & 14.93 & 63.69 & 49.82 \\
    &       & Rep-GoogLeNet  & 21.69 & 87.50 & 17.95 & 81.73 & 11.49 & 63.62 & 47.33 \\ \hline
\end{tabular}
\label{tab:abl-all}
\end{table*}

\begin{figure*}[t!]
    \centering
    \subfloat[Nano+NX]{
	\includegraphics[width=5.85cm,height=3.4cm]{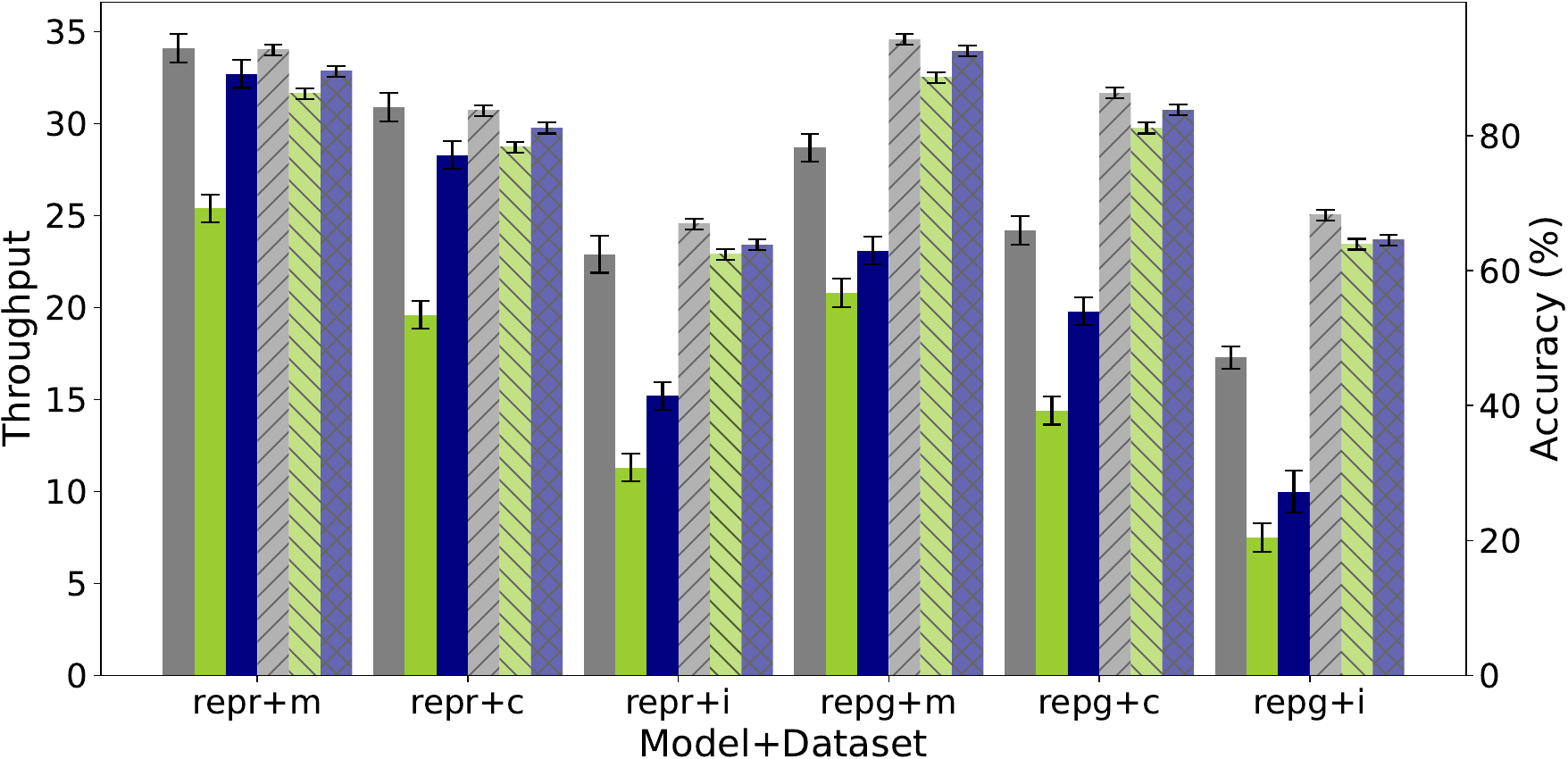}}
 \subfloat[TX1+NX]{
	\includegraphics[width=5.85cm,height=3.4cm]{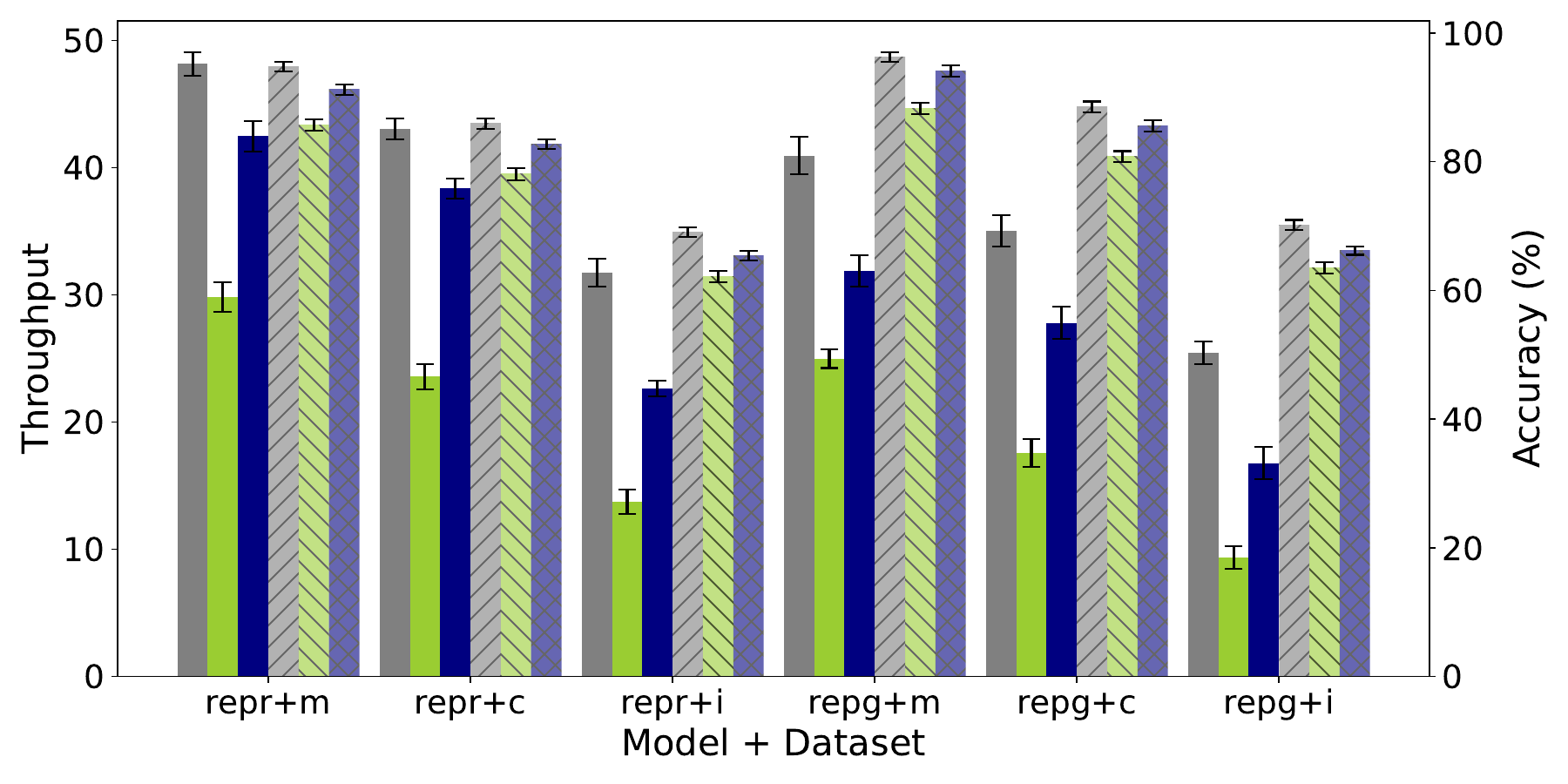}}
 \subfloat[TX2+NX]{
	\includegraphics[width=5.85cm,height=3.4cm]{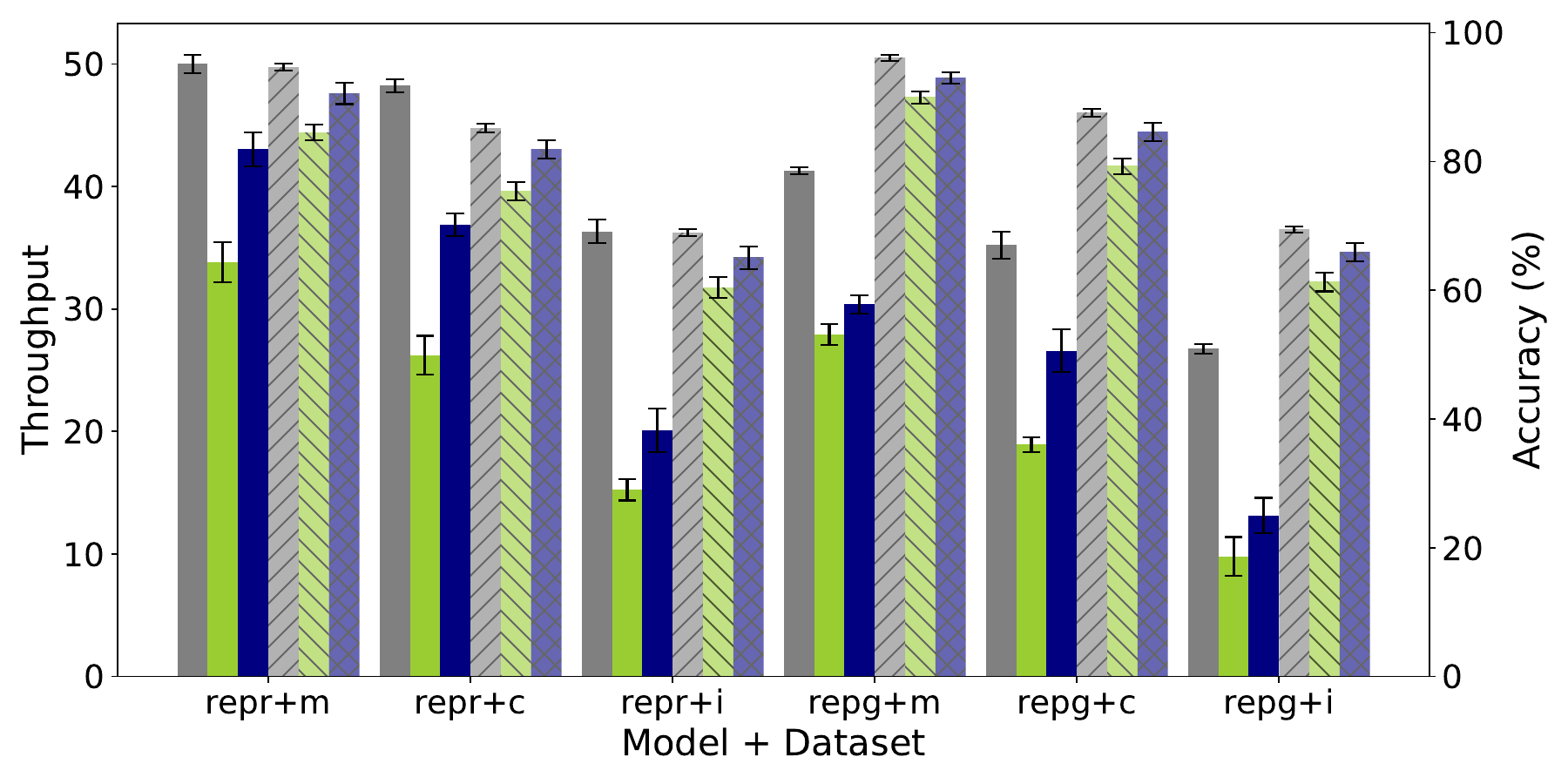}}

  \subfloat{
	\includegraphics[width=16cm,height=0.53cm]{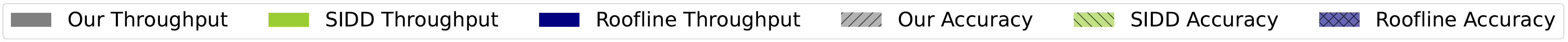}}
    \caption{The generalization experiment results are presented across different computing devices. The error bars in the bar charts represent data variance. “\textbf{repr}” denotes Rep-Resnet50, and “\textbf{repg}” refers to Rep-GoogLeNet, while "\textbf{m}", "\textbf{c}", and "\textbf{i}" represent the MNIST, CIFAR-100, and ImageNet datasets, respectively. The experimental results demonstrate that our method outperforms typical partition deployment strategies across different devices, exhibiting excellent generalizability.
}
\label{fig:abl-generdevice}
\end{figure*}

As shown in TABLE \ref{tab:abl-all}, the \textit{Generalizability Evaluation} results show that our method consistently outperforms two typical distributed deployment methods, SIDD and RoofSplit, across all datasets and base models. For instance, in the MNIST dataset, our method achieves the highest throughput of 43.17 images/second and accuracy of 92.47\%, compared to SIDD (32.13 images/second, 86.17\%) and RoofSplit (36.93 images/second, 88.25\%). This trend holds for CIFAR-100 and ImageNet as well, demonstrating the superior generalization ability of our method across diverse models and datasets. In the \textit{Improvements Evaluation}, when \textit{the roofline strategy} is removed, our method’s performance decreases significantly. On the MNIST dataset, removing the roofline strategy reduces the throughput from 43.17 to 41.72 images/second and accuracy from 92.47\% to 89.04\%. Similarly, for ImageNet, throughput drops from 22.64 to 19.36 images/second, highlighting that the roofline strategy plays a significant role in improving both throughput and accuracy. Without \textit{the dynamic model structure}, our method experiences a decline in performance, especially in throughput. On the CIFAR-100 dataset, throughput drops from 37.89 to 33.15 images/second, while accuracy decreases from 83.32\% to 79.67\%. This also results in a large decrease in accuracy. Additionally, we will conduct a detailed analysis in the "Case Analysis" section to examine the impact of the dynamic model structure on model accuracy and perceptual performance. The absence of \textit{the multi-objective function} leads to the most significant performance drop, especially in the accuracy metric. For instance, on the ImageNet dataset, removing this improvement reduces accuracy from 66.81\% to 62.35\%, and throughput from 22.64 to 16.04 images/second. These results underscore the importance of the multi-objective function in balancing the trade-offs between model partitioning and accuracy, ensuring optimal deployment performance.

The \textit{Average} results further emphasize the overall effectiveness of our approach, demonstrating that our method achieves the optimal balance between throughput and accuracy across various base models, making it highly suitable for IoVT system edge deployments.

\begin{figure}[t]
    \centering
    \includegraphics[width=\linewidth]{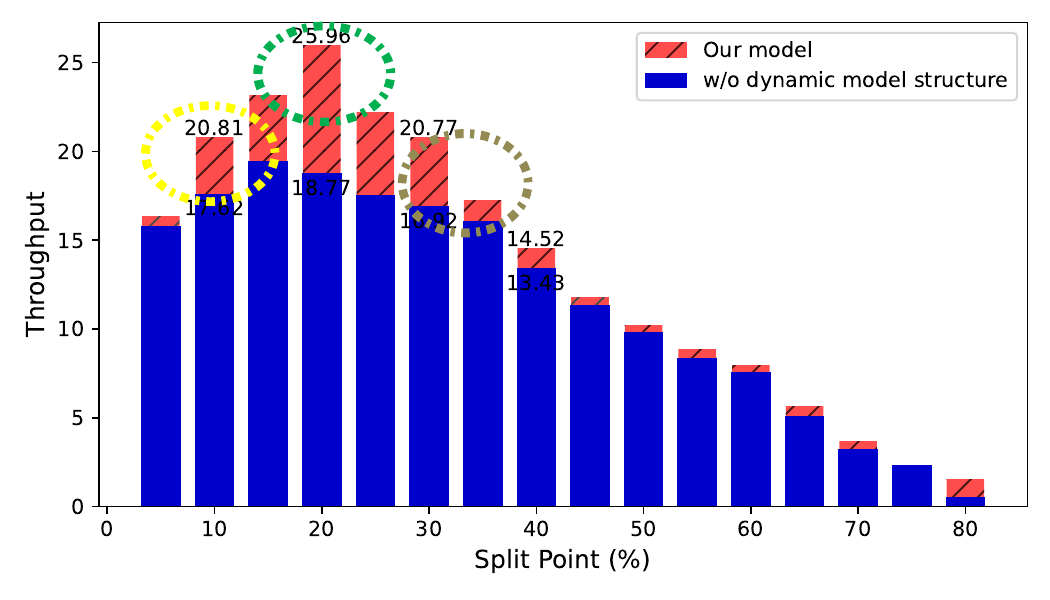}
    \caption{The throughput results are obtained under different split points on CIFAR-100. In this dynamic structure ablation experiment, the model is in the initial random structure, and the model structure is referred to as $\textit{S}_3$, as described in the section "\textbf{The Design of Dynamic Model Structure.}" The base model is set to Rep-GoogLeNet. Additionally, the different \textbf{split point} represents dividing the model into two parts  at varying proportions, where "\textbf{N\%}" indicates that N\% of the model's structure is deployed on Device 1, and "1-N\%" is deployed on Device 2. The \textbf{different colored dashed lines} represent the throughput at typical split points. Without the dynamic model structure, the throughput remains significantly lower than that of original method.}

\label{fig:abl-splitpoint}
\end{figure}

Considering the focus of this paper on practical applications within IoVT systems, the \textit{generalizability} of the method across different device is a key indicator of its effectiveness. We utilized Nano, TX1, and TX2 as computing terminals (Device 1) paired with the more powerful NX (edge server, Device 2) to execute edge inference tasks. As shown in Fig. \ref{fig:abl-generdevice}, based on the same base model, our method consistently achieves higher throughput and accuracy compared to SIDD and RoofSplit across all datasets. In particular, it significantly surpasses both typical partition deployment methods in terms of throughput. These results further demonstrate the excellent generalizability of our method. Unlike device-specific models designed by NAS\cite{li2023dlw}, it leverages co-optimization to adapt the base model to different devices, greatly enhancing its practical application value.

Furthermore, we find that the dynamic model structure has a significant impact, as shown in TABLE \ref{tab:abl-all}. Therefore, we further analyze the throughput across different split points when the dynamic model structure is removed, as illustrated in Fig. \ref{fig:abl-splitpoint}. The different split point represents the percentage of the model deployed on the computing terminal (Device 1, Nvidia Jetson Nano).  The trend of throughput shows an initial increase followed by a decrease, with the maximum value occurring at low-percentage split points. This indicates that maximum throughput is achieved when the majority of the model is deployed on the more powerful the edge server (Device 2, NVIDIA Jetson Xavier NX), which ensures the full utilization of both devices' computational capabilities—this aligns with common sense. Clearly, when the dynamic model structure is removed, the throughput is inferior to the original method, regardless of the split point chosen. Next, we analyze three typical split points in detail. The \textit{green dashed line} shows the point (around 15\%) at which throughput reaches its maximum. In this case, the dynamic model structure increases the computational complexity of sub-model 2 (the model deployed on Device 2) to fully leverage its computational power. The \textit{yellow dashed line} demonstrates that, even when the split point is located in the shallow layers of the model, the dynamic model structure can increase the computational complexity of sub-model 1 (the model deployed on Device 1) to maximize Device 1's performance. Similarly, the \textit{gray dashed line} shows that, even when sub-model 2 occupies a smaller proportion, the dynamic model structure can increase its computational complexity to further maximize Device 2’s computational potential and improve overall throughput. These results further visually illustrate the effectiveness of the dynamic model structure.

\subsection{Case Analysis: Coal Mine Helmet Detection Simulation Experiment}
Considering that the model will be deployed in IoVT systems for visual analysis, we focus on the task of object detection, a highly demanded function in vision applications. Therefore, we conduct additional object detection simulation experiments based on an IoVT system, using a coal mine video surveillance system as a real-world application scenario. The detailed experimental setup is as follows. We use NVIDIA Jetson TX2 as the embedded device and a computer with an RTX 4090 GPU as the server. TABLE \ref{tab:simulsetup} outlines the experimental setup specifications. Additionally, we set the communication conditions with an uplink bandwidth of 107.4 Mbps and a Ping latency of 13 ms. We present a coal mine object detection dataset (CMODD) based on images from the Ordos Mataitou coal mine surveillance system. The dataset contains 12,000 images with a resolution of 640 × 400 and is publicly available at \href{https://github.com/word-ky/Co-design-Sys-NN}{\textit{https://github.com/word-ky/Co-design-CMODD-Access}}.

\begin{table}[h!]
\setlength{\tabcolsep}{0.03cm}
\centering
\caption{Experimental Devices \& Network Environment.}
{\begin{tabular}{cc}
\hline
\multirow{5}{*}{\textbf{NVIDIA Jetson TX2}}         & Peak Computational Power: 1.33 TFLOPS \\
                                           & Memory Bandwidth: 59.7 GB/s \\
                                           & GPU Model: 256-core NVIDIA \\
                                           & Pascal™ architecture GPU \\
                                           & Storage: 32GB eMMC 5.1
                                           \\ \hline
\multirow{5}{*}{\textbf{NVIDIA RTX 4090}}           & Peak Computational Power: 82.6 TFLOPS \\
                                           & Memory Bandwidth: 1018 GB/s \\
                                           & GPU Model: 16,384-core NVIDIA \\
                                           & Ada Lovelace architecture GPU \\
                                           & VRAM: 24GB GDDR6X \\ \hline
\multirow{4}{*}{\textbf{Network Environment}}                         & Uplink bandwidth: 107.4 Mbps \\
                                                      & Downlink bandwidth: 48.7 Mbps \\
                                                        & Packet loss: 0\% \\
                                                       & Ping latency: 13 ms \\ \hline
\end{tabular}}
\label{tab:simulsetup}
\end{table}

\begin{table}[t!]
    \centering
    \setlength{\tabcolsep}{0.15cm} 
    \caption{The backbone based on dynamic network
structure. }
    \begin{tabular}{lcccc}
        \toprule
        \textbf{Index} & \textbf{Module} & \textbf{Output Channels} & \textbf{Kernel Size} & \textbf{Stride} \\
        \midrule
        0 & Conv & 16 & 3*3 & 2 \\
        \midrule
        1 & Conv & 32 & 3*3 & 2 \\
        \midrule
        \rowcolor[gray]{0.9} 2-0 & RepBlock & 32 & 1*1, 3*3 & 1 \\
        \rowcolor[gray]{0.9} 2-1 & RepBlock & 32 & 1*1, 3*3 & 1 \\
        \rowcolor[gray]{0.9} 2-2 & RepBlock & 32 & 1*1, 3*3 & 1 \\
        \midrule
        3 & Conv & 64 & 3*3 & 2 \\
        \midrule
        \rowcolor[gray]{0.9} 4-0 & RepBlock & 64 & 1*1, 3*3 & 1 \\
        \rowcolor[gray]{0.9} 4-1 & RepBlock & 64 & 1*1, 3*3 & 1 \\
        \rowcolor[gray]{0.9} 4-2 & RepBlock & 64 & 1*1, 3*3 & 1 \\
        \rowcolor[gray]{0.9} 4-3 & RepBlock & 64 & 1*1, 3*3 & 1 \\
        \rowcolor[gray]{0.9} 4-4 & RepBlock & 64 & 1*1, 3*3 & 1 \\
        \rowcolor[gray]{0.9} 4-5 & RepBlock & 64 & 1*1, 3*3 & 1 \\
        \midrule
        5 & Conv & 128 & 3*3 & 2 \\
        \midrule
        \rowcolor[gray]{0.9} 6-0 & RepBlock & 128 & 1*1, 3*3 & 1 \\
        \rowcolor[gray]{0.9} 6-1 & RepBlock & 128 & 1*1, 3*3 & 1 \\
        \rowcolor[gray]{0.9} 6-2 & RepBlock & 128 & 1*1, 3*3 & 1 \\
        \rowcolor[gray]{0.9} 6-3 & RepBlock & 128 & 1*1, 3*3 & 1 \\
        \rowcolor[gray]{0.9} 6-4 & RepBlock & 128 & 1*1, 3*3 & 1 \\
        \rowcolor[gray]{0.9} 6-5 & RepBlock & 128 & 1*1, 3*3 & 1 \\
        \midrule
        7 & Conv & 256 & 3*3 & 2 \\
        \midrule
        \rowcolor[gray]{0.9} 8-0 & RepBlock & 256 & 1*1, 3*3 & 1 \\
        \rowcolor[gray]{0.9} 8-1 & RepBlock & 256 & 1*1, 3*3 & 1 \\
        \rowcolor[gray]{0.9} 8-2 & RepBlock & 256 & 1*1, 3*3 & 1 \\
        \midrule
        9 & SPPF & 256 & 5*5 & 1 \\
        \bottomrule
    \end{tabular}
    \label{tab:simulbackbone_structure}
\end{table}

\begin{figure*} [b!]
	\centering

    \subfloat[]{
		\includegraphics[width=3cm]{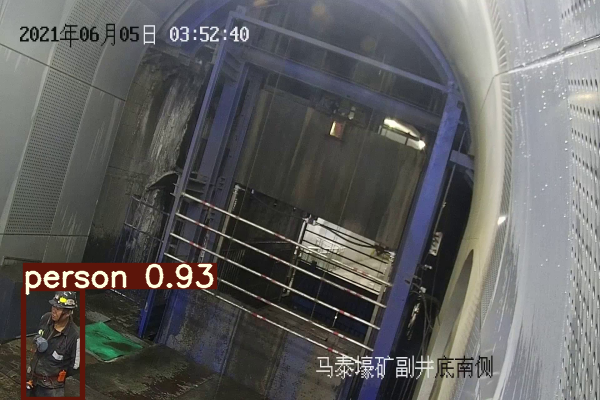}}
    \hspace{-3mm}
    \subfloat[]{
		\includegraphics[width=3cm]{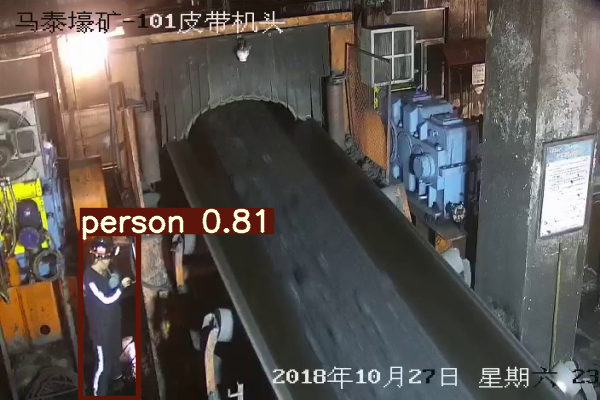}}
    \hspace{-3mm}
    \subfloat[]{
		\includegraphics[width=3cm]{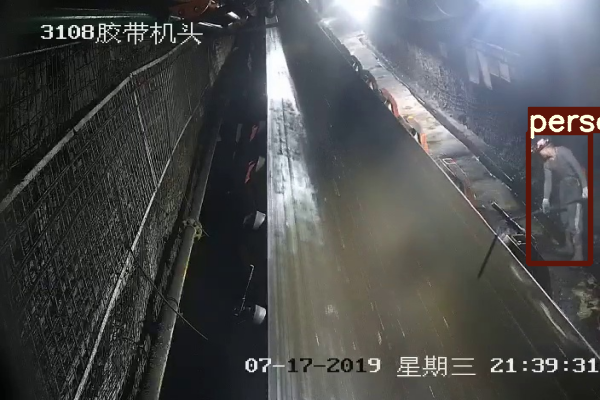}}
    \hspace{-3mm}
    \subfloat[]{
		\includegraphics[width=3cm]{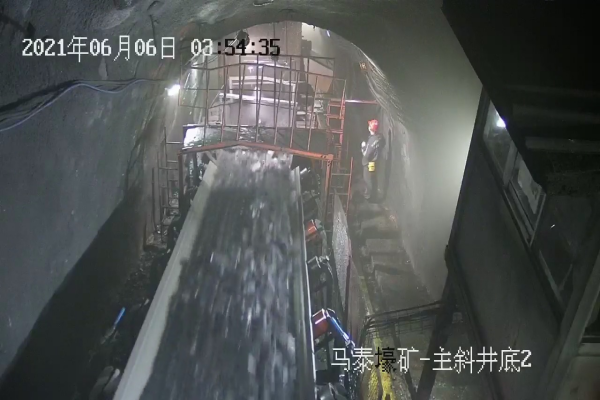}}
    \hspace{-3mm}
    \subfloat[]{
		\includegraphics[width=3cm]{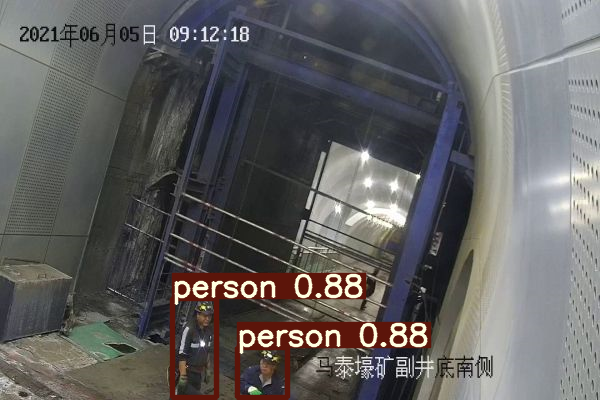}}
    \hspace{-3mm}
    \subfloat[]{
		\includegraphics[width=3cm]{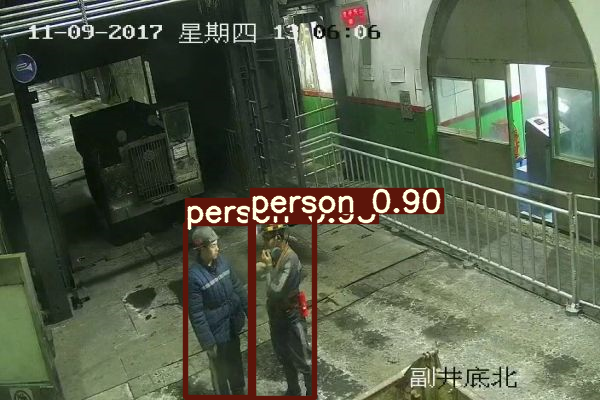}}
    \hspace{-3mm}
    
    \setcounter{subfigure}{0}      
    \centering{RoofSplit}
  
    \subfloat[]{
		\includegraphics[width=3cm]{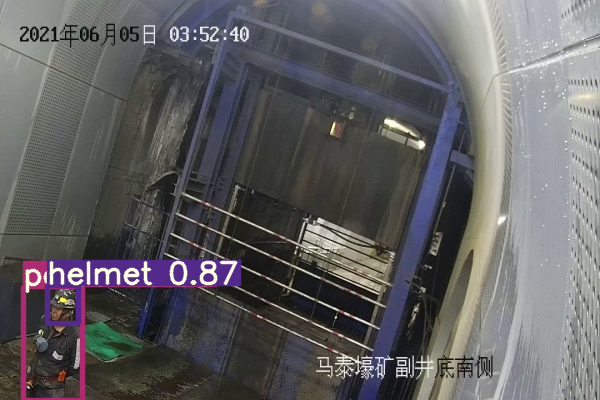}}
    \hspace{-3mm}
    \subfloat[]{
		\includegraphics[width=3cm]{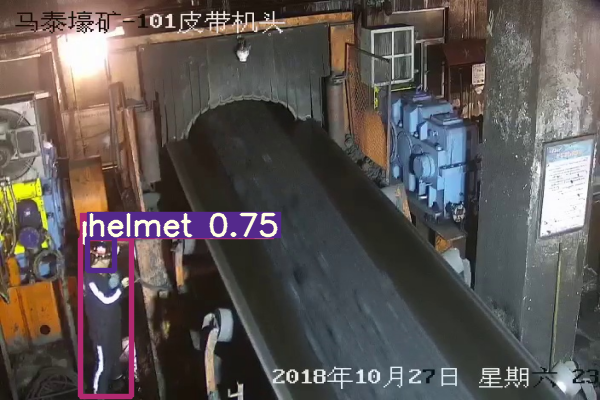}}
    \hspace{-3mm}
    \subfloat[]{
		\includegraphics[width=3cm]{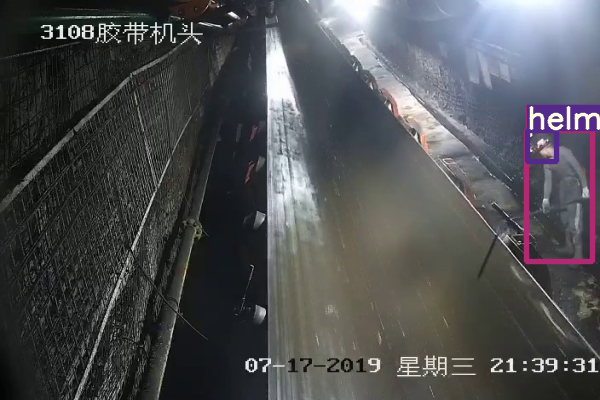}}
    \hspace{-3mm}
    \subfloat[]{
		\includegraphics[width=3cm]{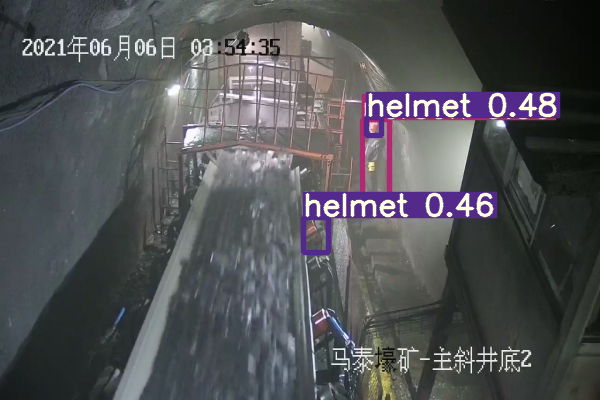}}
    \hspace{-3mm}
    \subfloat[]{
		\includegraphics[width=3cm]{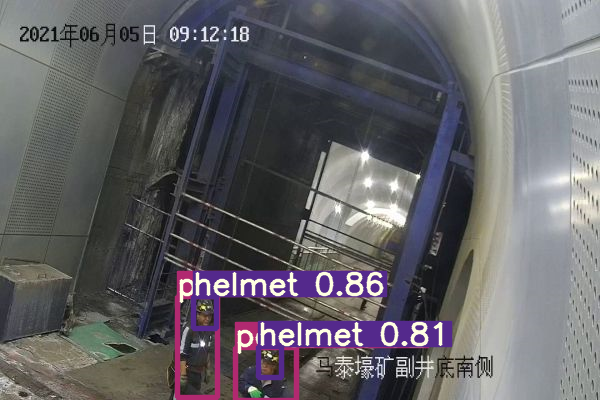}}
    \hspace{-3mm}
    \subfloat[]{
		\includegraphics[width=3cm]{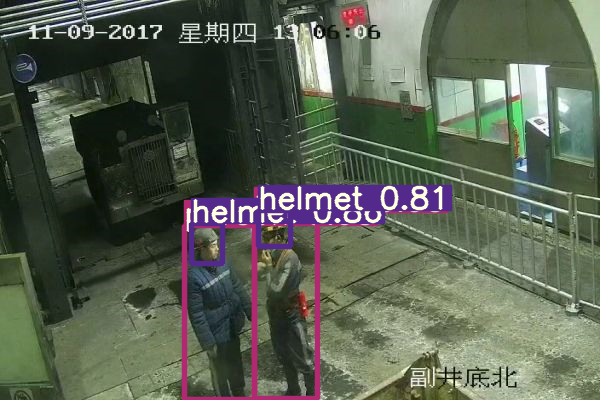}}
    \hspace{-3mm}
    
    \setcounter{subfigure}{0}      
    \centering{Our method}
  
	\caption{The detection results of Roofsplit and our method. RoofSplit is capable of detecting large-sized objects such as '\textbf{person},' but consistently fails to detect small-sized objects like '\textbf{helmet}.' In contrast, while our method still exhibits false positives in certain cases, such as in \textbf{(d)}, it achieves accurate detection of small-sized objects in the majority of scenarios.}
\label{fig:simul-comparisonvisual}
\end{figure*}

\begin{table}[t]
\centering
\setlength{\tabcolsep}{0.02cm} 
\caption{Performance Comparison of Different Methods. Our method demonstrates \textbf{competitive detection accuracy} and achieves the \textbf{highest throughput}, significantly surpassing the second-best, RoofSplit. In terms of the overall performance metric, \textbf{Average}, it achieved the highest score. These results demonstrate that it achieves a \textbf{balance} between detection accuracy and throughput on the edge of IoVT systems, showcasing its \textbf{practical applicability}.}
\begin{tabular}{ccccc}
\toprule
                                           \textbf{Method} &  \textbf{Acc-helmet} &  \textbf{Acc-person} &  \textbf{Throughput} &  \textbf{Average} \\
\midrule
                                             SIDD\cite{sidd} &       \textcolor{red}{84.27} &       \textcolor{red}{90.19} &       17.41 &    \textcolor{blue}{63.96} \\
                                        RoofSplit\cite{roofsplit} &       71.93 &       84.06 &       21.34 &    59.11 \\
                                       Our method &       79.26 &       86.12 &       \textcolor{red}{27.98} &    \textcolor{red}{64.45} \\
\makecell{Our method with \\ Cloud-edge computing approach} &       80.49 &       87.07 &       12.38 &    59.98 \\
     \makecell{Our method with \\Cloud computing approach} &       \textcolor{blue}{82.15} &       \textcolor{blue}{88.69} &        12.20 &    61.01 \\
\bottomrule
\end{tabular}
\label{tab:simula-comparison}
\end{table}

Moreover, we choose the widely-used YOLO family\cite{zou2023object} models as the base models, which are extensively applied in industrial applications\cite{sidd, wu2024lightweight, zhao2024small}. The YOLO model framework typically consists of Backbone, Neck, and Head modules, where the backbone is the key component for feature extraction and encoding\cite{terven2023comprehensive}. Based on prior experience\cite{li2018detnet,jiang2022review}, the primary parameters and computational complexity of YOLO models reside in the backbone. Additionally, given that the computational resources of edge servers generally exceed those of the terminal devices, our analysis indicates that the partition point must lie within the backbone, a conclusion supported by previous experimental results, as shown in Fig. \ref{fig:abl-splitpoint}. To align with practical applications and avoid unnecessary complexity, we limit our analysis on the impact of the dynamic network structure of the backbone. The following TABLE \ref{tab:simulbackbone_structure} shows the dynamic backbone structure built on the re-parameterization strategy, and we use the RepVGG, as shown in Fig. \ref{fig:dynamicmodelstructure}, as the RepBlock. Moreover, we employ the Neck and Head modules of YOLOv8s as the other components in the YOLO framework\cite{terven2023comprehensive}. Detection accuracy is measured using \textit{mAP@IoU}=0.5, and we use the \textit{Acc-X} to represent the accuracy for specific categories. For instance, '\textit{Acc-helmet}' is used to reflect the model's performance in detecting small-sized objects, while real-time performance is evaluated by the \textit{throughput}.

We evaluate the performance differences between our method and four comparison methods, including two typical algorithms: RoofSplit\cite{roofsplit} and SIDD\cite{sidd}, as well as two variations of our method based on the cloud computing approach: the total cloud computing approach (where the model is fully deployed on cloud servers) and the cloud-edge computing approach (where the end device is removed, and the model is split between the cloud server and the edge server).

\textit{Comparison analysis on accuracy and throughput.} SIDD\cite{sidd} achieves high detection accuracy by leveraging a super-resolution strategy, which enhances feature perception. However, this approach significantly impacts its throughput. RoofSplit\cite{roofsplit} can be considered a variation of our method without the dynamic model structure. Given the importance of the dynamic model structure, RoofSplit’s detection accuracy is understandably lower than that of our method. For example, RoofSplit shows a 7.3\% decrease in accuracy for the helmet detection category compared to our approach. Furthermore, since both the ``\textit{cloud computing}'' and ``\textit{cloud-edge computing}'' methods do not use TensorRT acceleration on cloud servers, they exhibit slightly improved accuracy compared to the original method. Specifically, helmet detection accuracy improves by approximately 2.4\% in the cloud computing version (82.15 vs. 79.26) and by 1.6\% in the cloud-edge computing version (80.49 vs. 79.26). However, both cloud-based methods introduce an additional data transmission phase from the edge to the cloud, which significantly reduces system throughput. For instance, the throughput in the cloud computing version drops to 12.2, a substantial reduction compared to 27.98 in the edge-based version---representing a 56.4\% decrease in throughput. Furthermore, as the volume of transmitted data increases, edge-cloud transmission delays grow exponentially, severely affecting system response time. In terms of overall performance, our method achieves the highest \textit{Average} score of 64.45, effectively balancing detection accuracy and throughput. This result underscores the practical value of our method, particularly in scenarios where both high accuracy and real-time performance are essential.

Small object detection has significant practical application demands, making this performance metric an important criterion for evaluating detection models. We visualize the detection results between RoofSplit\cite{roofsplit} and our method to intuitively show the difference in small object detection, as shown in Fig. \ref{fig:simul-comparisonvisual}. Unlike our method, RoofSplit fails to detect small-sized objects, such as 'helmet,' and struggles to identify objects with fine details (e.g., in (d)). These results clearly demonstrate that our method is capable of accurate small object detection.

\begin{figure} [h!]
	\centering	
        \subfloat[]{
		\includegraphics[width=1in]{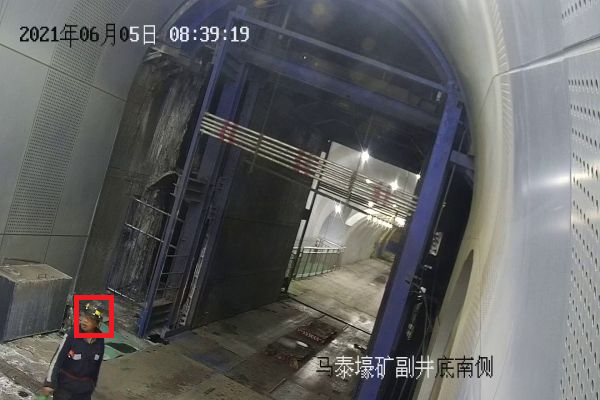}}
        \subfloat[]{
		\includegraphics[width=1in]{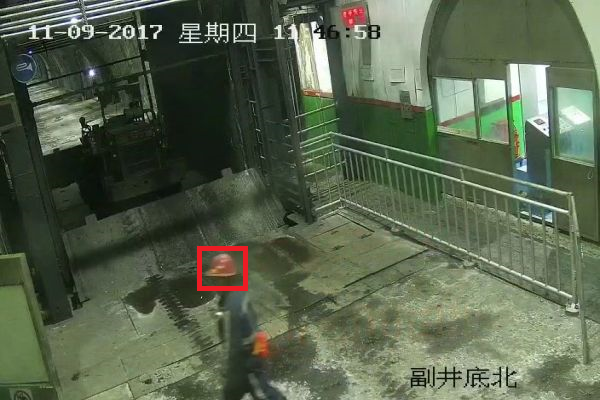}}
        \subfloat[]{
		\includegraphics[width=1in]{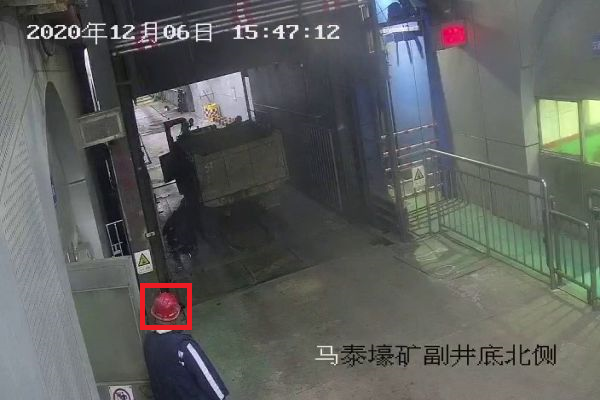}}
  \setcounter{subfigure}{0}      
    \centering{Original Images}
	          
        \subfloat[]{
		\includegraphics[width=1in]{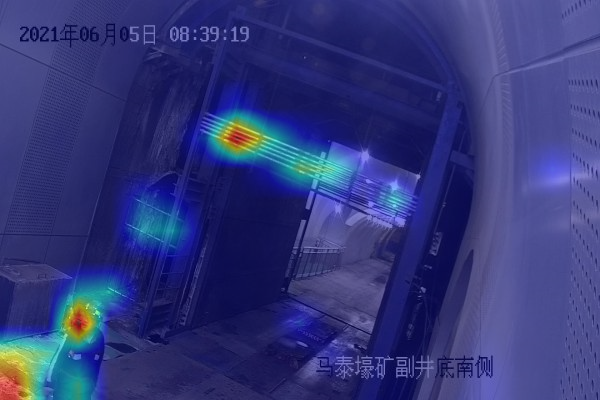}}
        \subfloat[]{
		\includegraphics[width=1in]{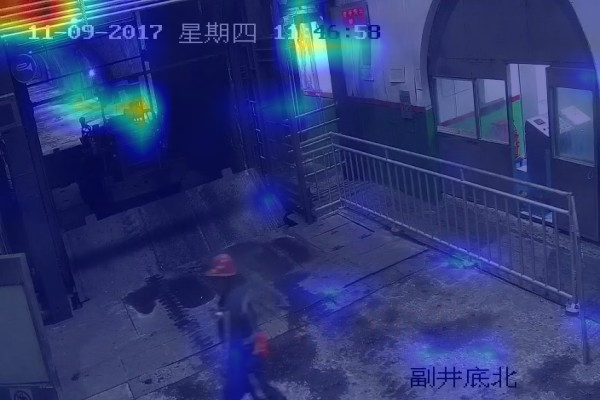}}
        \subfloat[]{
		\includegraphics[width=1in]{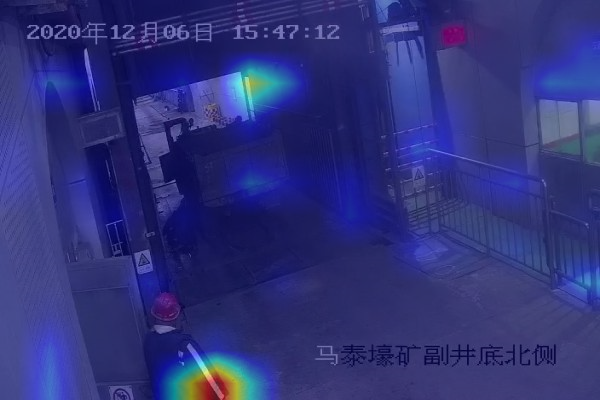}}
\setcounter{subfigure}{0}

    \centering{Our method w/o dynamic model structure}
    
        \subfloat[]{
		\includegraphics[width=1in]{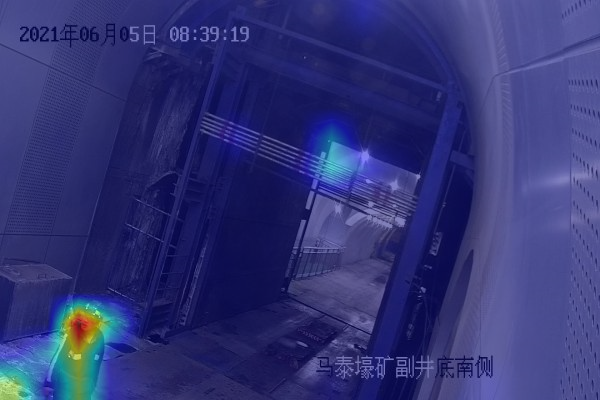}}	
        \subfloat[]{
		\includegraphics[width=1in]{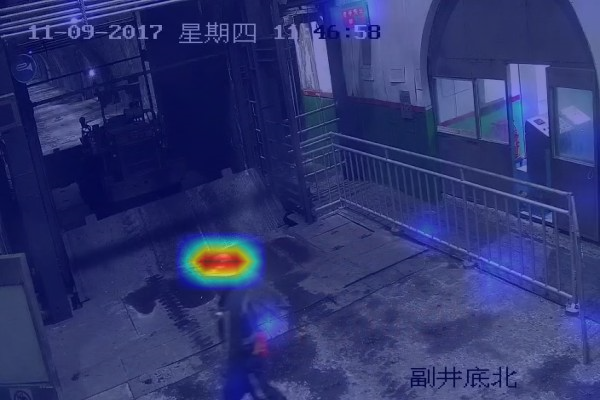}}
        \subfloat[]{
		\includegraphics[width=1in]{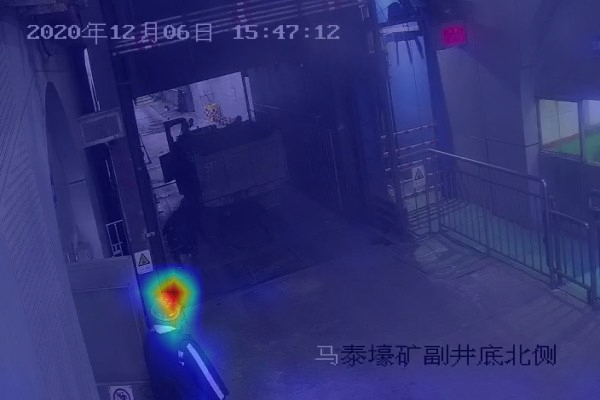}}
  \setcounter{subfigure}{0}

    \centering{Our method}
    
	\caption{The comparison of perception results on small-size objects. Removing the dynamic model structure significantly impairs our method's ability to detect small objects. In contrast, with the dynamic model structure, our method accurately captures small-sized objects.}
\label{fig:simulpercp}
\end{figure}

\textit{The analysis of small-size object detection performance.} Compared to our method, RoofSplit\cite{roofsplit} can be regarded as the base model with the $\textit{S}_3$ structure, rather than the dynamic model structure. We use heatmaps to further highlight the small-object detection capability of our method. When the dynamic model structure is removed, detection errors such as false positives, as seen in Fig. \ref{fig:simulpercp} (a), and missed detections, as observed in (b) and (c), become more frequent. This indicates a weakened ability to perceive small objects. In contrast, with the support of the dynamic model structure, our method accurately locates small, specific targets, enabling precise detection. This further illustrates our method’s superior capability in detecting small objects with the support of a dynamic model structure.

\begin{figure} [h!]
	\centering	
        \subfloat[Original Image]{
		\includegraphics[width=1in]{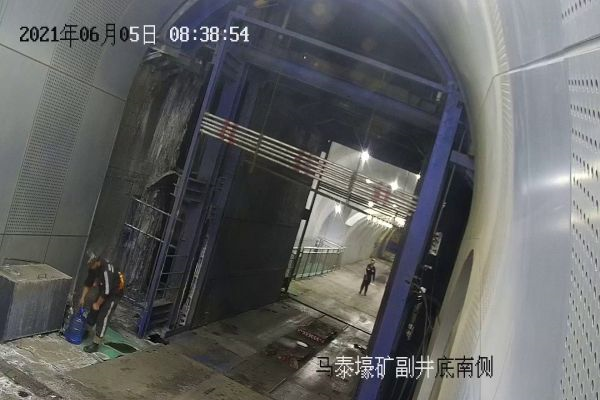}}
        \subfloat[Base model with $\textit{S}_3$]{
		\includegraphics[width=1in]{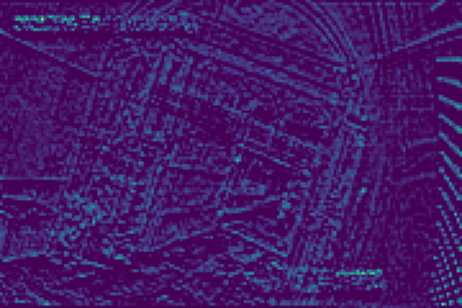}}
        \subfloat[Our method]{
		\includegraphics[width=1in]{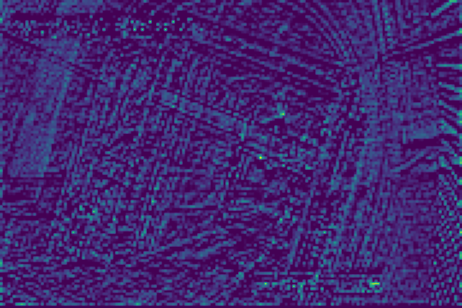}}
  \setcounter{subfigure}{0}      
    \centering{Example I}
	          
        \subfloat[Original Image]{
		\includegraphics[width=1in]{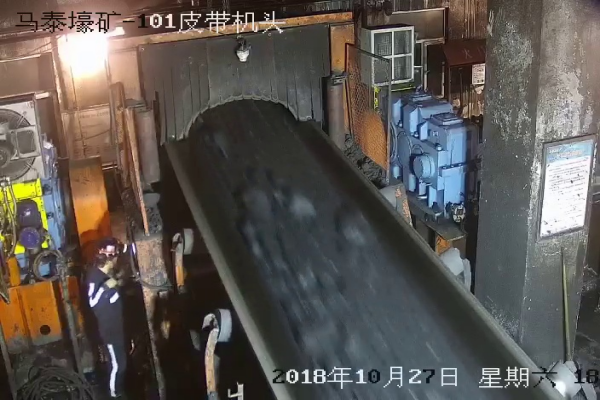}}
        \subfloat[Base model with $\textit{S}_3$]{
		\includegraphics[width=1in]{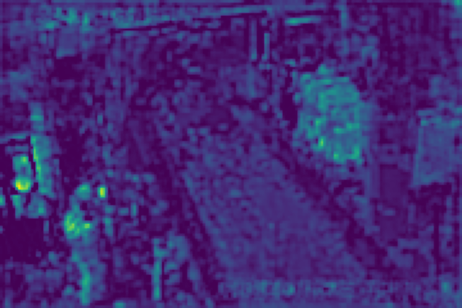}}
        \subfloat[Our method]{
		\includegraphics[width=1in]{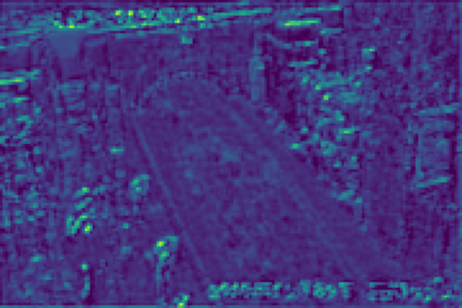}}
\setcounter{subfigure}{0}      
    \centering{Example II}

	\caption{Comparison of feature extraction capabilities. If the backbone block of the base model, as shown in TABLE \ref{tab:simulbackbone_structure}, uses only the '$\textit{S}_3$' structure, high-level features such as contours and edges can be extracted. In contrast, a backbone with the dynamic model structure is able to extract more detailed features, which enhances small-object detection performance.}
\label{fig:simulfeatex}
\end{figure}

Furthermore, we analyze the small-object detection performance of our method from the perspective of feature extraction, as shown in Fig. \ref{fig:simulfeatex}. The base model equipped with $\textit{S}_3$ (i.e., the fused 3*3 convolutional layer) can only capture large-scale features, such as edges and contours. By comparison, when the dynamic model structure shifts to $\textit{S}_3$ + $\textit{S}_1$ + $\textit{S}_s$, it supplements the model with additional small-sized convolutional structures, enabling the base model to extract a richer set of fine-grained features. For network architectures that incorporate multiple convolutional sizes, such as GoogLeNet, which includes 7*7, 5*5, 3*3, and 1*1 convolutional channels, the performance gap caused by the fusion strategy becomes more pronounced. This explains why our dynamic model structure enhances the classification accuracy of Rep-GoogLeNet, as shown in TABLE \ref{tab:abl-all}. In our method, under the multi-objective optimization framework, the dynamic model structure adaptively adds $\textit{S}_3$ + $\textit{S}_1$ + $\textit{S}_s$ to the base model, improving the ability to capture detailed features and thereby enhancing the performance in small-object detection.

\section{Conclusion $\&$ Future Work} 

We propose a \textbf{Co-design Framework of Neural Networks and Edge Deployment} to realize real-time visual inference for IoVT systems. Specifically, we utilize Roofline analysis to perform partitioned model deployment, and we introduce a re-parameterization-based dynamic model structure to further enhance model-device alignment, maximizing computational performance to increase throughput. Additionally, we employ a multi-objective co-optimization approach to balance throughput and accuracy in the partitioned deployment and network fusion. Extensive experiments demonstrate that, compared to benchmark algorithms, our method achieves higher throughput and superior classification accuracy. Furthermore, it has the advantage of generalizability, as it leverages the co-optimization approach to adapt pre-trained models across different devices. This applies well to practical IoVT systems with varying device registration and deregistration scenarios. Additional detection simulation results show that our method enables real-time and accurate detection at the edge of IoVT systems in real-world industrial scenarios, especially for small-scale objects.

Despite the proposed co-design method’s ability to achieve high throughput and accuracy, along with generalizability and practical application value, some limitations remain:
\begin{itemize}
    \item \textbf{Dependency on Specific Network Structures:} The dynamic model structure relies on the re-parameterization principle for channel fusion with different convolutional kernels. While many neural networks use multi-channel structures, this approach is not universally compatible with all network architectures, significantly restricting its application scope.

    \item \textbf{Limited Throughput Improvement:} Traditional Roofline analysis primarily focuses on the relationship between compute intensity and memory bandwidth, often overlooking dynamic factors such as resource allocation and workload balancing during model-device adaptation. This limitation restricts Roofline analysis from fully evaluating throughput potential across different devices.
\end{itemize}

To address the above issues, \textbf{firstly}, we draw inspiration from low-rank optimization methods \cite{hu2021lora, yu2023compressing, gu2024mix} and incorporate rank optimization into the design of dynamic model structures, thus extending beyond traditional multi-channel convolutions to a broader range of network architectures. Rank optimization enables the model to maintain computational efficiency and accuracy while adapting to diverse neural network structures, such as shallow networks or those with fewer convolutional channels. This approach significantly reduces the model's dependency on specific network configurations (e.g., multi-channel or fixed kernel setups), thereby greatly enhancing its generalizability. \textbf{Secondly}, we will refine the traditional Roofline analysis by integrating dynamic resource allocation and workload balancing mechanisms on the device. This enhancement aims to optimize model-device alignment and, ultimately, increase throughput.



\begin{IEEEbiography}[{\includegraphics[width=1in,height=1.25in,clip,keepaspectratio]{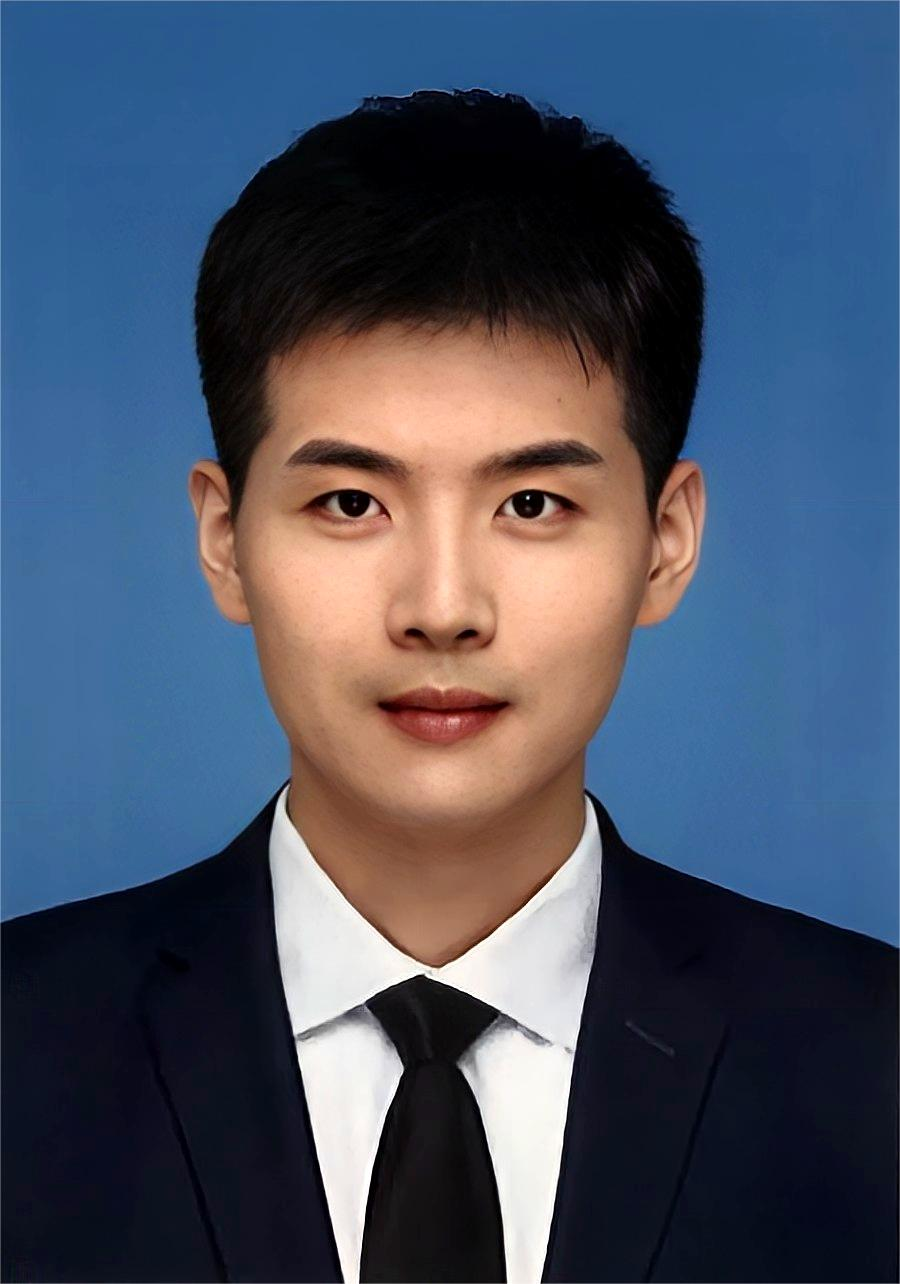}}]{Jiaqi Wu}
is a third-year doctoral candidate at China University of Mining and Technology(Beijing), and his research focuses on computer vision and computer graphics. He was sent by Chinese Scholarship Council to the University of British Columbia as a joint doctoral student in 2023 to study the Internet of Things and blockchain technology.\end{IEEEbiography}

\begin{IEEEbiography}[{\includegraphics[width=1in,height=1.25in,clip,keepaspectratio]{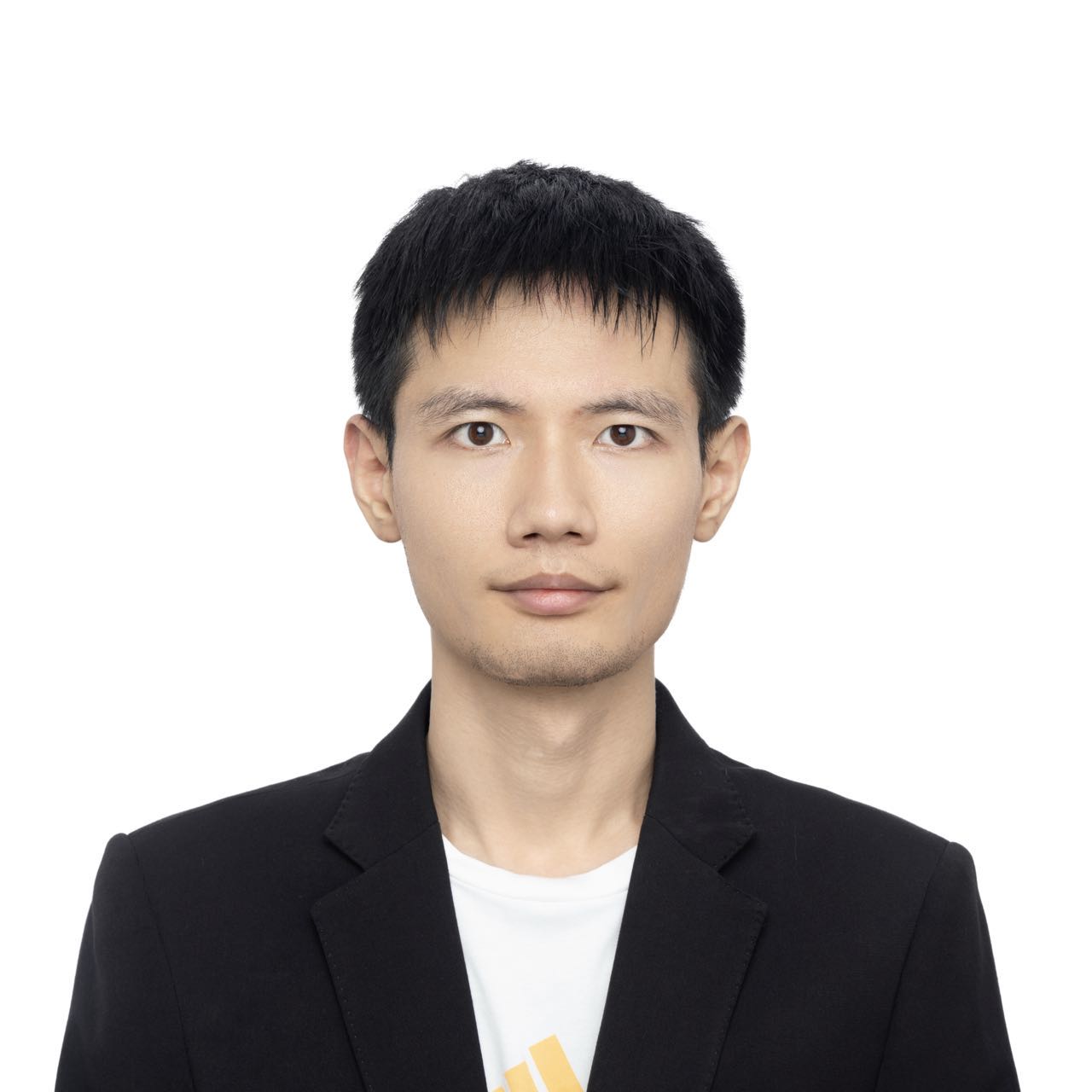}}]{Simin Chen}
received his Ph.D. in Computer Science from the University of Texas at Dallas, USA, in 2024. Prior to this, he earned his Master's degree from Tongji University in May 2018. Afterward, he joined Columbia University as a postdoctoral researcher. His research interests span machine learning, software engineering, and program analysis.\end{IEEEbiography}

\begin{IEEEbiography}[{\includegraphics[width=1in,height=1.25in,clip,keepaspectratio]{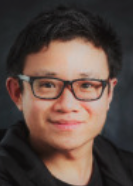}}]{Zehua Wang}{
received Ph.D. degree from The University of British Columbia (UBC), Vancouver in 2016. He is an Adjunct Professor in the Department of Electrical and Computer Engineering at UBC, Vancouver and the CTO at Intellium Technology Inc. He implemented the payment system for commercialized mobile ad-hoc networks with the payment/state channel technology on blockchain. His is interested in protocol and mechanism design with optimization and game theories to improve the efficiency and robustness for communication networks, distributed systems, and social networks.}\end{IEEEbiography}

\begin{IEEEbiography}[{\includegraphics[width=1in,height=1.25in,clip,keepaspectratio]{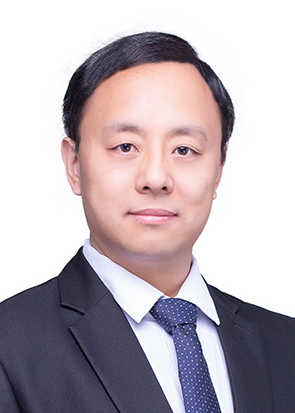}}]{Wei Chen}
[M’18] received the Ph.D. degree in communications and information systems from China University of Mining and Technology, Beijing, China, in 2008. In 2008, he joined the School of Computer Science and Technology, China University of Mining and Technology, where he is currently a professor. His research interests include machine learning, image processing, and computer networks.\end{IEEEbiography}

\begin{IEEEbiography}[{\includegraphics[width=1in,height=1.25in,clip,keepaspectratio]{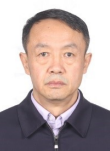}}]{Zijian Tian}
received the Ph.D. degree in communications and information systems from China University of Mining and Technology, Beijing, China, in 2003. In 2003, he joined the School of Mechanical Electronic and Information Engineering, China University of Mining and Technology(Beijing), where he is currently a professor. He is a member of National Expert Group of Safety Production, and is a member and convener of Expert Committee of Information and Automation. His research interests include image processing, analysis of coal mine surveillance video.\end{IEEEbiography}

\begin{IEEEbiography}[{\includegraphics[width=1in,height=1.25in,clip,keepaspectratio]{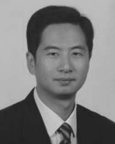}}]{F. Richard Yu}
\textcolor{black}{received the PhD degree in electrical engineering from the University of British Columbia (UBC) in 2003. His research interests include connected/autonomous vehicles, artificial intelligence, blockchain, and wireless systems. He has been named in the Clarivate’s list of "Highly Cited Researchers" in computer science since 2019, Standford’s Top 2$\%$ Most Highly Cited Scientist since 2020. He received several Best Paper Awards from some first-tier conferences, Carleton Research Achievement Awards in 2012 and 2021, and the Ontario Early Researcher Award (formerly Premiers Research Excellence Award) in 2011. He is a Board Member the IEEE VTS and the Editor-in-Chief for IEEE VTS Mobile World newsletter. He is a Fellow of the IEEE, Canadian Academy of Engineering (CAE), Engineering Institute of Canada (EIC), and IET. He is a Distinguished Lecturer of IEEE in both VTS and ComSoc.}\end{IEEEbiography}

\begin{IEEEbiography}[{\includegraphics[width=1in,height=1.25in,clip,keepaspectratio]{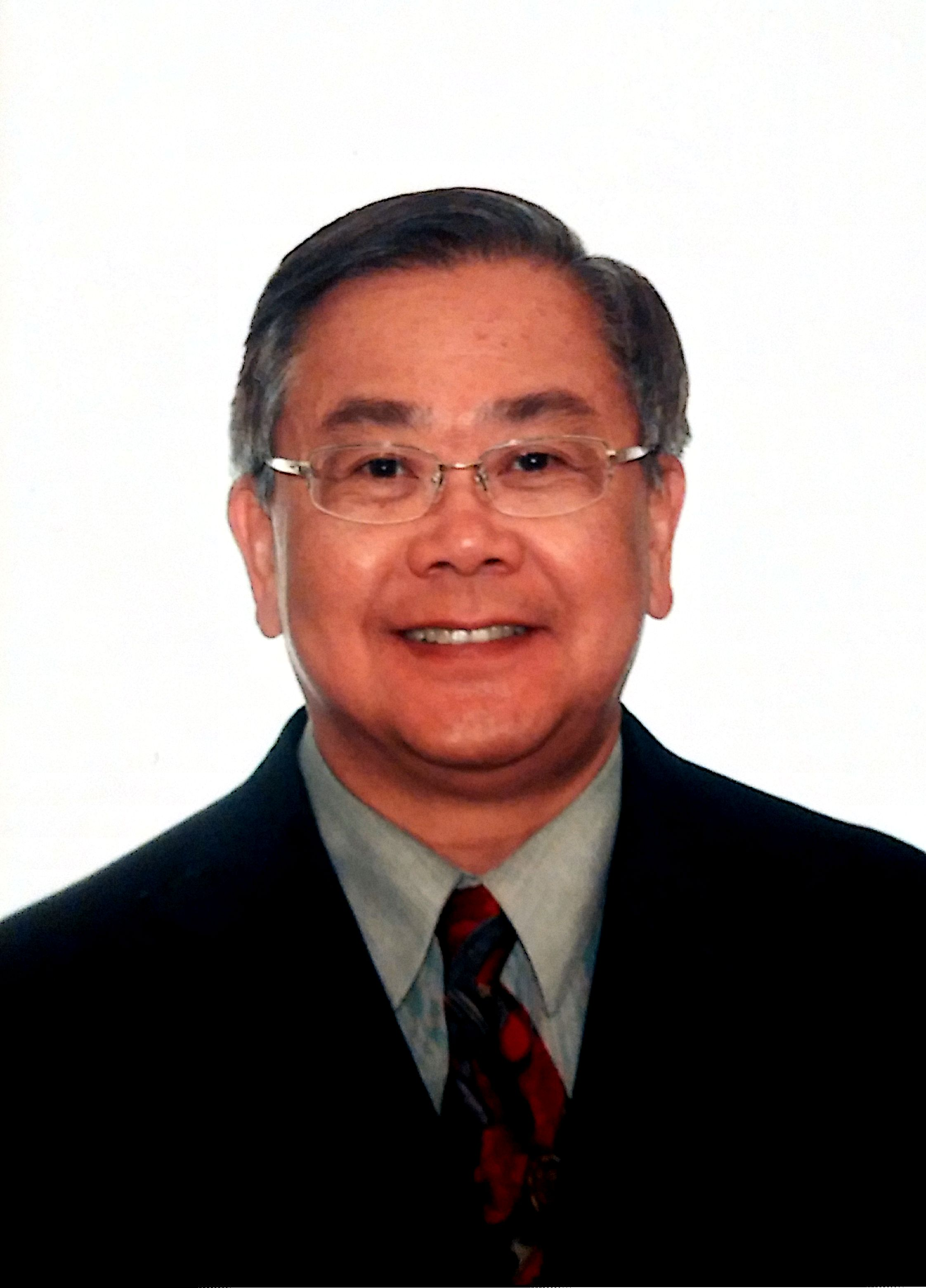}}]{VICTOR C. M. LEUNG}{ (Life Fellow, IEEE) is the Dean of the Artificial Intelligence Research Institute at Shenzhen MSU-BIT University (SMBU), China, a Distinguished Professor of Computer Science and Software Engineering at Shenzhen University, China, and also an Emeritus Professor of Electrical and Computer Engineering and Director of the Laboratory for Wireless Networks and Mobile Systems at the University of British Columbia (UBC), Canada.  His research is in the broad areas of wireless networks and mobile systems, and he has published widely in these areas. His published works have together attracted more than 60,000 citations. He is named in the current Clarivate Analytics list of “Highly Cited Researchers”. Dr. Leung is serving on the editorial boards of the IEEE Transactions on Green Communications and Networking, IEEE Transactions on Computational Social Systems, and several other journals. He co-authored papers that were selected for the 2017 IEEE ComSoc Fred W. Ellersick Prize, 2017 IEEE Systems Journal Best Paper Award, 2018 IEEE CSIM Best Journal Paper Award, and 2019 IEEE TCGCC Best Journal Paper Award.  He is a Life Fellow of IEEE, and a Fellow of the Royal Society of Canada (Academy of Science), Canadian Academy of Engineering, and Engineering Institute of Canada. }\end{IEEEbiography}

\end{document}